\documentclass[10pt,twocolumn,letterpaper]{article}

\usepackage{cvpr}
\usepackage{times}
\usepackage{epsfig}
\usepackage{graphicx}
\usepackage{subfigure}
\usepackage[ruled,linesnumbered]{algorithm2e}
\usepackage{amsmath}
\usepackage{amssymb}

\cvprfinalcopy % *** Uncomment this line for the final submission

\def\cvprPaperID{} % *** Enter the CVPR Paper ID here

\begin{document}

%%%%%%%%% TITLE
\title{Clustering via Boundary Erosion}

\author{Cheng-Hao Deng\\
Computer Science Department\\
Xiamen University\\
{\tt\small chenghaodeng@stu.xmu.edu.cn}
% For a paper whose authors are all at the same institution,
% omit the following lines up until the closing ``}''.
% Additional authors and addresses can be added with ``\and'',
% just like the second author.
% To save space, use either the email address or home page, not both
\and
Wan-Lei Zhao\thanks{Corresponding author. Fujian Key Laboratory of Sensing and Computing for Smart City, Xiamen University, Fujian, China.}\\
Computer Science Department\\
Xiamen University\\
{\tt\small wlzhao@xmu.edu.cn}}

\maketitle
%\thispagestyle{empty}
% this must go after the closing bracket ] following \twocolumn[ ...

% This command actually creates the footnote in the first column
% listing the affiliations and the copyright notice.
% The command takes one argument, which is text to display at the start of the footnote.
% The \icmlEqualContribution command is standard text for equal contribution.
% Remove it (just {}) if you do not need this facility.

%\printAffiliationsAndNotice{the corresponding author}  % leave blank if no need to mention equal contribution
%\printAffiliationsAndNotice{\icmlEqualContribution} % otherwise use the standard text.

\begin{abstract}
Clustering analysis identifies samples as groups based on either their mutual closeness or homogeneity. In order
to detect clusters in arbitrary shapes, a novel and generic solution based on boundary erosion is proposed. The clusters are assumed to be separated by relatively sparse regions. The samples are eroded sequentially according to their dynamic boundary densities. The erosion starts from low density regions, invading inwards, until all the samples are eroded out. By this manner, boundaries between different clusters become more and more apparent. It therefore offers a natural and powerful way to separate the clusters when the boundaries between them are hard to be drawn at once. With the sequential order of being eroded, the sequential boundary levels are produced, following which the clusters in arbitrary shapes are automatically reconstructed. As demonstrated across various clustering tasks, it is able to outperform most of the state-of-the-art algorithms and its performance is nearly perfect in some scenarios.
\end{abstract}

\section{Introduction}

Clustering problems arise from variety of applications, such as documents/web pages categorization~\cite{ml04:zhao}, pattern recognition, biomedical analysis~\cite{nature15}, data compression via vector quantization~\cite{SiZ03} and nearest neighbor search~\cite{JDS11,pami14:flann}. In general, clustering analysis plays an indispensable role for understanding various phenomena across different contexts. Given a set of samples $S$ in \textit{d}-dimensional space $R^d$, the task of clustering is to partition the data samples into subsets (called clusters) such that the samples in the same cluster are more homogeneous or closer to each other than those from different subsets (clusters). 

Traditionally, this issue has been modeled as a distortion minimization problem in \textit{k}-means~\cite{km82}. The clustering procedure is organized into two steps. Firstly, samples are assigned to their closest centers. Secondly, the center of each cluster is updated with the data samples assigned to it. These two steps are repeated until the structure of clusters does not change in two consecutive iterations. This algorithm is simple and efficient whereas it is unable to discover clusters that are not in spherical shape. 

Aiming to identify clusters of arbitrary shapes, a series of algorithms have been proposed in the last two decades. Most of these algorithms~\cite{dbscan,meanshift,sci14:alex, density_review, optics99, decode09, rank_order} are conceived from the perspective of density distribution. Intuitively, samples within each cluster are concentrated, and clusters are separated by sparse regions. Among these algorithms, samples are either iteratively assigned to~\cite{dbscan,optics99,sci14:alex} or shifted towards the density peaks~\cite{meanshift}. The clusters are therefore forged. However, heuristic rules~\cite{dbscan} or kernels~\cite{meanshift} that are employed in the algorithms are unable to deal with various density distributions in practice. For this reason, the performance of these algorithms turns out to be unstable under different scenarios.

Apart from the density based approaches, graph based algorithms are also able to discover clusters of arbitrary shapes. Representative methods are Chameleon~\cite{chameleon99} and order-constrained transitive distance clustering~\cite{ocdt16:zdyu}. In both of them, the connectivity between samples are carefully considered. According to the strategies presented in the papers, samples that are far away from each other are still clustered together as long as they are reachable to each other via a chain of closely connected bridging samples. Unfortunately, for both of them, a matrix that keeps pair-wise distance between samples is required, which makes it inscalable to large-scale clustering tasks.

As a consequence, despite numerous efforts have been taken in the last several decades, two major goals in clustering analysis, namely the ability of identifying clusters in arbitrary shapes and the scalability towards large scale and high dimensional data, are hardly achieved with one algorithm. In this paper, a simple but effective density based solution is proposed. The basic idea is inspired by the phenomenon of land erosion by water. The boundaries between clusters are drawn gradually by a boundary erosion process without any heuristic rules or kernels. This is particularly powerful in the case that boundaries between clusters are obscure at the first sight. In addition, the bit-by-bit boundary erosion produces a sequential order following which the potential clusters could be reconstructed with the guidance of an \textit{r}-NN graph. The boundary erosion is feasible in any metric spaces as long as the density of data sample could be estimated. Furthermore, we also demonstrate that this algorithm achieves satisfactory performance and high efficiency on large-scale and high dimensional clustering task with the support of efficient \textit{k}-NN graph construction~\cite{weidong}.

\section{Related Work}
\label{sec:rela}
Since the proposal of \textit{k}-means, a variety of clustering algorithms have been proposed in the past three decades, which are in general categorized into seven groups~\cite{itnn05:xu}. Namely, they are agglomerative~\cite{agnes90}, divisive, partitioning~\cite{km82}, density based~\cite{dbscan,decode09,optics99,sci14:alex,dbscan15,meanshift}, graph based~\cite{chameleon99} and neural network based algorithms. In the literature, it is also seen the ensemble of several existing algorithms to boost the performance~\cite{ensemble12:zhou,ensemble14:zheng}. For the comprehensive surveys, readers are referred to~\cite{itnn05:xu,ijirr13:greenlaw}. In this section, our focus will be on the review of several typical algorithms that are able to identify clusters in arbitrary shapes. Namely, the density based and graph based algorithms are discussed mainly.

Although clustering problem has been modeled from different perspectives, people basically agree clusters are composed by samples that are relatively concentrated and are separated in-between by relatively sparse regions. This perception is made without any specification about the distance measure on the input data. Density based algorithms are designed in general in line with this perception. Although different in details, the density based algorithms aims to discover groups of samples that are continuously connected. 

In general, two steps are involved in the density based clustering process. Firstly, the local density surrounding each sample is estimated. Given sample $x_i$ and radius $r$, density of sample $x_i$ is defined as the number of samples ($y_i$s) that fall into $x_i$'s neighborhood of range $r$ (as shown in Eqn.~\ref{eqn:dns}). 

%\begin{eqnarray}
\begin{equation}
\rho(x_i) = \sum_{j}\sigma_j, \\
\mbox{where~}
\sigma_j =\left \{ \begin{array}{cc}
1 & d(x_i,y_j) \leq r \\ 
0 & \mbox{otherwise}
\end{array}. 
\right.
\label{eqn:dns}
%\end{eqnarray}
\end{equation}
Function $d(\cdot,\cdot)$ in Eqn.~\ref{eqn:dns} returns the distance between $x_i$ and $y_j$. In the second step, the clusters are forged basically in two different manners. For instance, in DBSCAN~\cite{dbscan}, cluster is formed by expanding it from ``core points'' (points hold high density) to points with low density. While in mean-shift clustering~\cite{meanshift}, data samples are shifted iteratively from region of low density towards the density peaks. In DBSCAN, the expansion process could be very sensitive to the parameters. For instance, two heterogeneous clusters are falsely merged into one as the parameter changes slightly. While in mean-shift, the shifting process could be easily stuck in a local if there is no obvious density peak. In the approach of clustering based on density peak (clusterDP)~\cite{sci14:alex}, data samples are directly assigned to the closest density peak in which each density peak is recognized as a cluster center. However, it faces similar problem as mean-shift since it is hard to identify the cluster center when there is no obvious density peak. Another pitfall of this approach is that the number of peaks to be selected as the cluster center has to be set manually.

%The graph based methods are also designed to discover clusters in arbitrary shapes. Representative algorithms are Chameleon~\cite{chameleon99} and Order-constrained transitive distance clustering (OCTD)~\cite{ocdt16:zdyu}. In Chameleon, the clustering procedure is organized into three steps. In the first step, the distance between every two samples is calculated, which therefore supplies the connectivity information between samples. In the second step, Chameleon uses a graph-partitioning algorithm to cluster the samples into several relatively small sub-clusters. During the third step, two intermediate sub-clusters that are mostly coupled are merged into one each time. The merging operation repeats until no sub-cluster pair satisfy with the merging condition. While for OCTD, similar as Chameleon, a distance matrix is required. The distance between sample pairs is revised based on the transitivity, which shortens the distance between samples if they are reachable to each other via a chain of closely connected samples. In the second step, the clustering is undertaken on the revised distance matrix in the similar way as spectral clustering~\cite{normcut}. Although encouraging performance is achieved, these algorithms face similar problems. First of all, it is computationally inefficient to calculate the full distance matrix. Furthermore, several extra parameters are introduced in both cases. It is not easy to fine-tune the parameters to achieve good performance when dealing with different types of data. 

Recently, several clustering methods are proprosed to deal with high dimensional data, which are hardly separable in the original space. Methods are typically in two categories. One is built upon generative model~\cite{dec,vade}. Another  is called sparse spectral clustering (SSC)~\cite{ssc,nips17:deepssc}. SSC projects the high-dimensional data into a sparse and low-dimensional representation first. The projected data are clustered via spectral clustering. Method in~\cite{iccv17} in general follows similar framework to fulfill the clustering on high-dimensional data.

\section{Clustering via Boundary Erosion}
\label{sec:alg}
\subsection{Motivation}
\begin{figure}
\begin{center}
	\subfigure[gradual erosion (step 1)]
	{\includegraphics[width=0.43\linewidth]{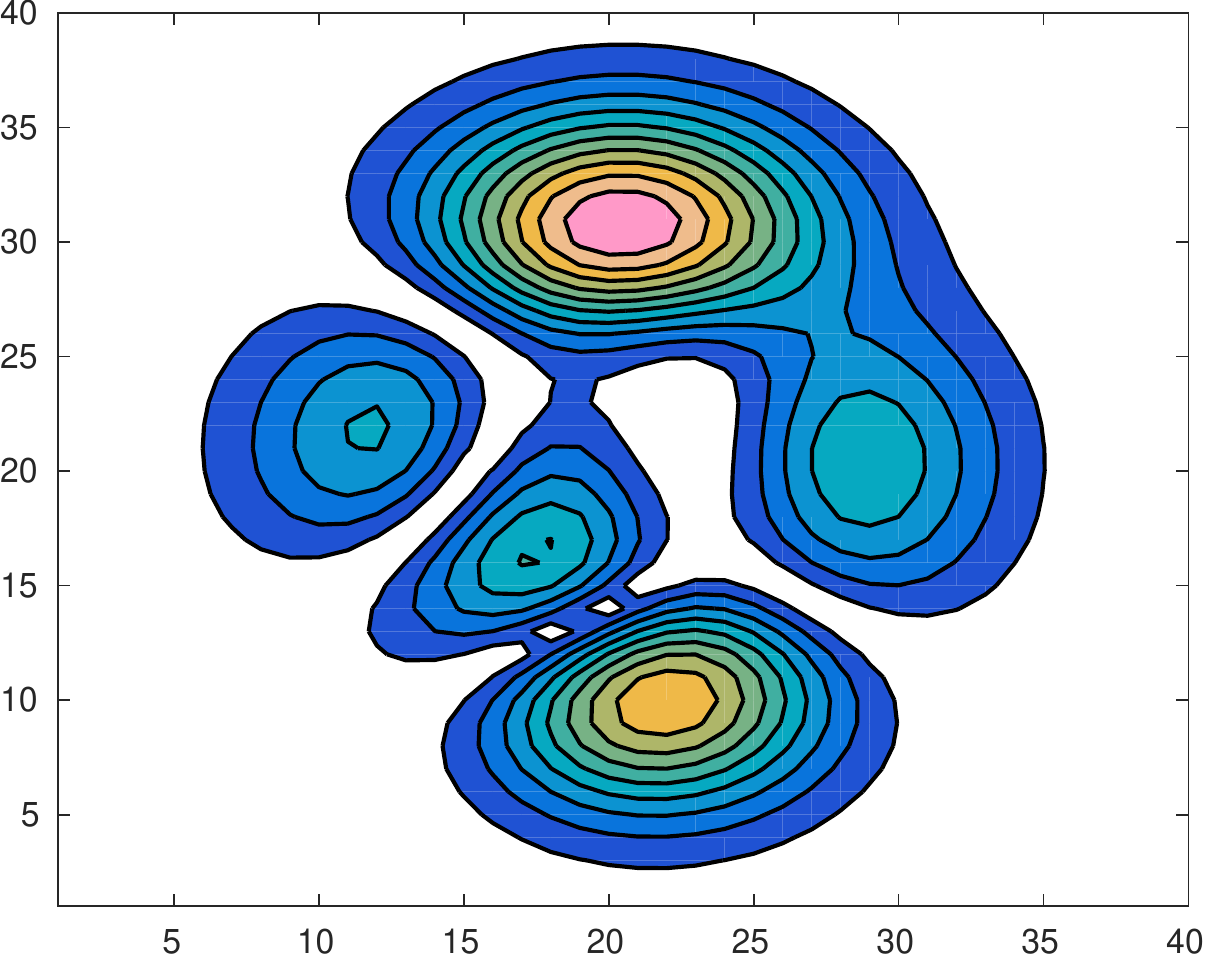}}
	\hspace{0.12in}
	\subfigure[gradual erosion (step 2)]
	{\includegraphics[width=0.43\linewidth]{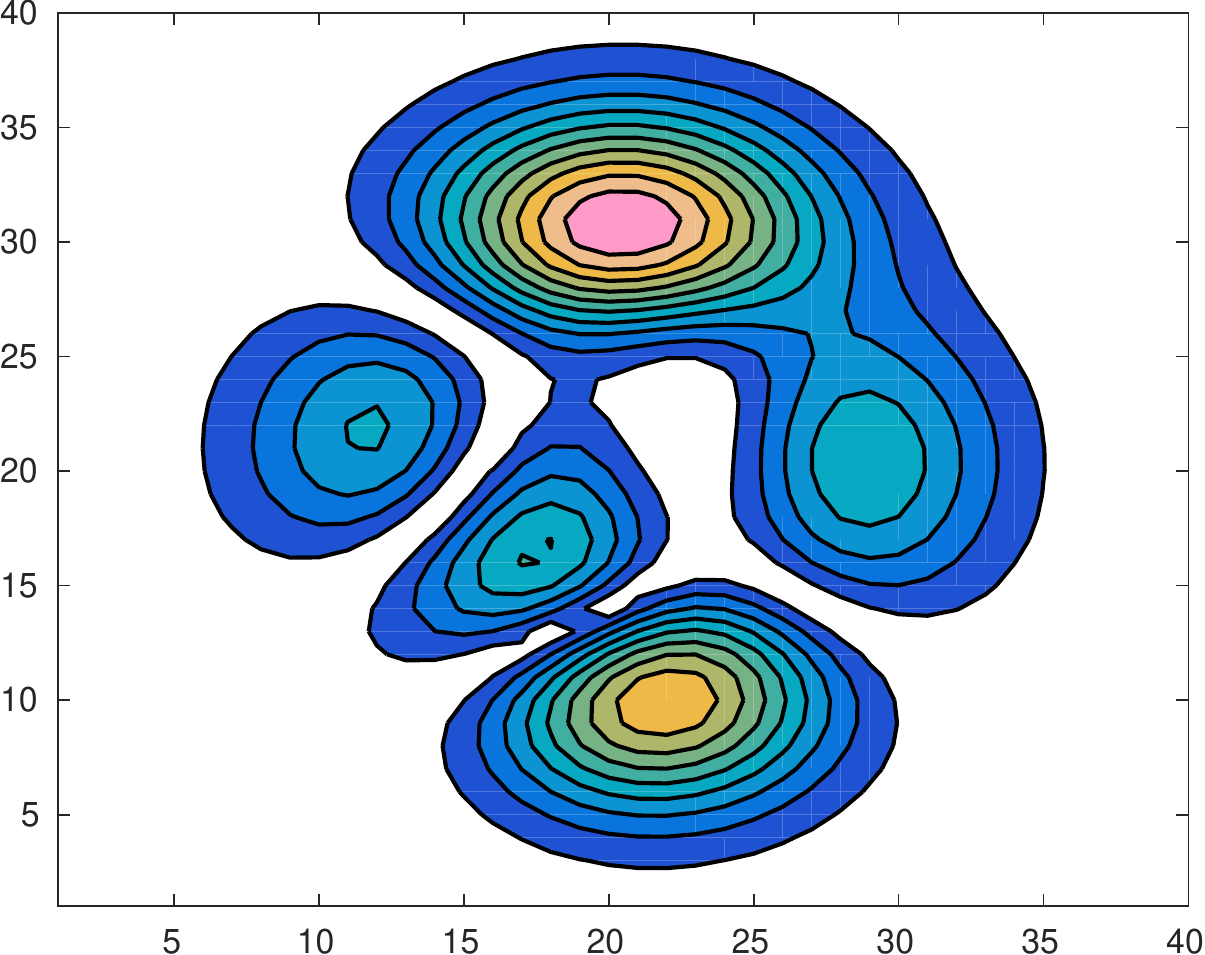}}\\
	\subfigure[gradual erosion (step 3)]
	{\includegraphics[width=0.43\linewidth]{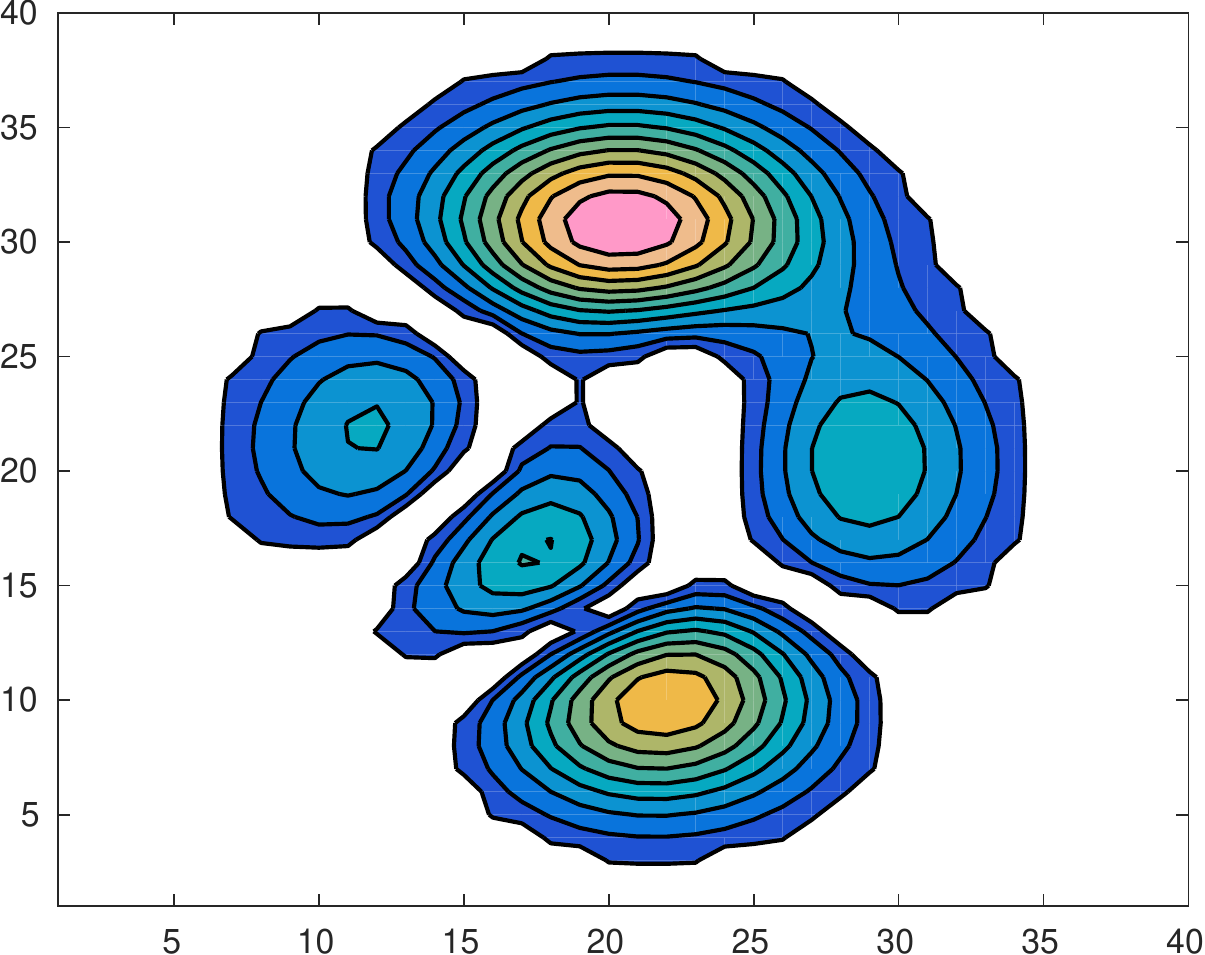}}
	\hspace{0.12in}
	\subfigure[trend of erosion]
	{\includegraphics[width=0.43\linewidth]{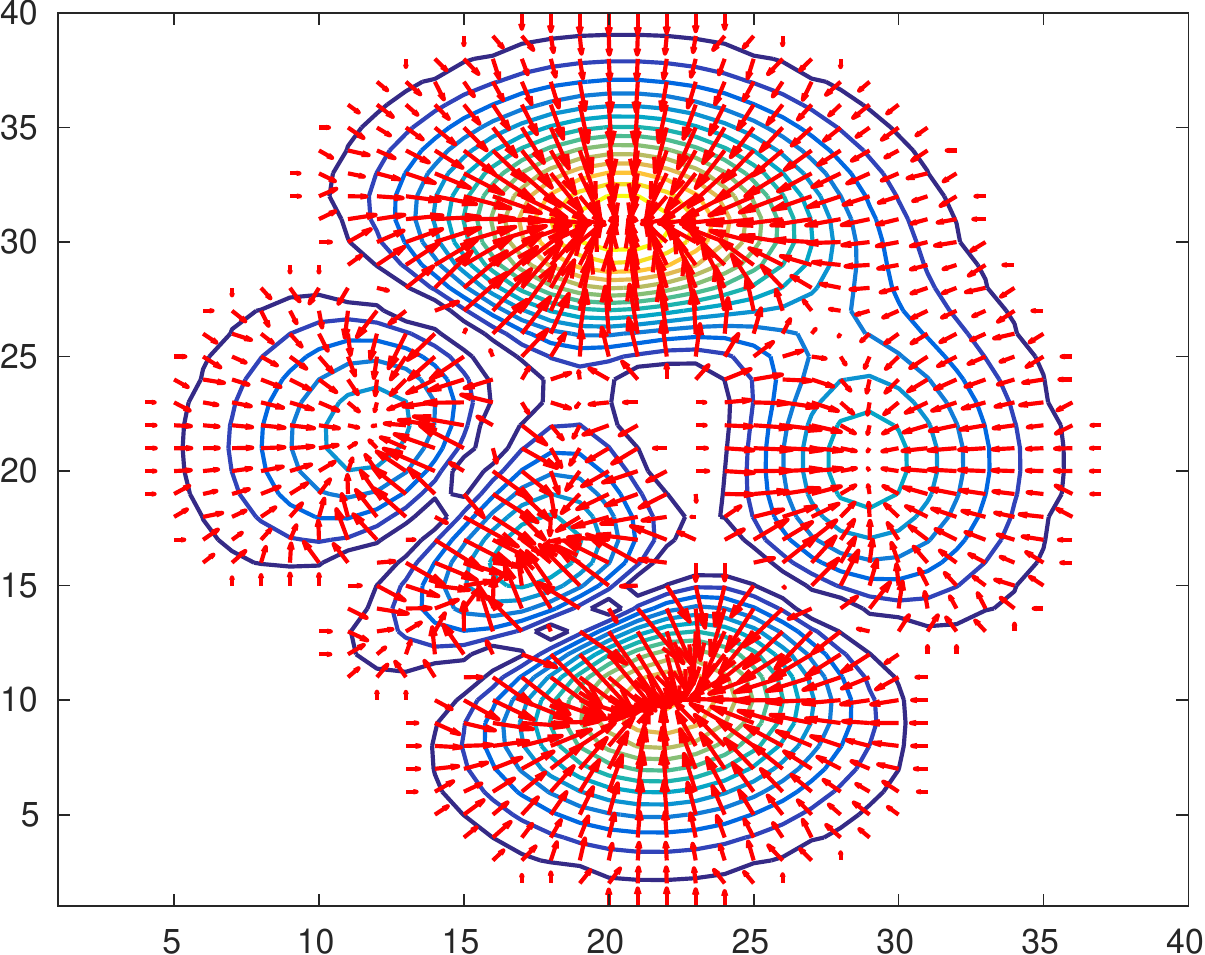}}
	\caption{An illustration of boundary erosion. The erosion starts from bottomlands, invading inwards. As more and more lands (from (a) to (c)) are eroded out, the boundaries between lands (clusters) become more and more apparent. The erosion continues until all the lands are eroded out.
\label{fig:bedemo}}
\vspace{-0.2in}
\end{center}
\end{figure}

For all the clustering algorithms discussed, the key step is to partition the data samples into groups. However, this is challenging particularly when boundaries between different clusters are not obvious. In this paper, a boundary erosion procedure is proposed, which addresses such kind of ambiguity. The idea is inspired by the natural phenomenon of land erosion by water. An illustration is given in Fig.~\ref{fig:bedemo}. As shown in the figure, the erosion explicitizes the boundaries between clusters gradually as the water erodes the lands bit by bit. More importantly, a sequential order that indicates how the samples are assembled one after another as one cluster is established based on the order of the samples being eroded. With this sequential order, which is called as sequential boundary levels in the paper, the latent clusters could be easily reconstructed.

Notice that this idea is essentially different from watershed transform~\cite{watershed00} in the sense that ``water level'' in our case does not rise up to bury the lands. Instead, it only erodes the lands. The lands on the outer part are eroded earlier than the inner lands instead of being buried at the same time, even they are on the same altitude.

\subsection{Generating Boundary Levels via Erosion}
\label{sec:belevel}
To facilitate the boundary erosion, the density of one sample is estimated in a quite different manner from conventional algorithms. Namely, the density of each sample is dynamically estimated by gradually eroding its neighbors out. In order to do that, a dynamic array $Q$, in which samples along with their dynamic boundary density $\rho^*$ is maintained. The dynamic boundary density is given in Eqn.~\ref{eqn:dnsp}.

%\begin{eqnarray}
\begin{equation}
\label{eqn:dnsp}
\rho^*(x_i) = \sum_{j}\sigma_j, \\ 
\mbox{where~}
\sigma_j =\left \{ \begin{array}{cc}
1 & d(x_i,y_j) \leq r \& j \in Q \\
0 & \mbox{otherwise}
\end{array}. 
\right.
\end{equation}
%\end{eqnarray}

As shown in Eqn.~\ref{eqn:dnsp}, the major difference from Eqn.~\ref{eqn:dns} is that samples outside of $Q$ are not counted during density estimation. At the beginning, all the samples are put into the dynamic array $Q$. For this reason, $\rho^*$ of each sample is initially the same as $\rho$ given in Eqn.~\ref{eqn:dns}. 

The boundary erosion starts from deleting the sample with the lowest density (which corresponds to boundaries we are most cerntain) in $Q$. Each time, data sample holding the lowest density\footnote{It is possible that several samples hold the same density value will be removed at once.} is removed out from $Q$. Due to the removal of sample $x_i$, dynamic boundary density $\rho^*$ of its neighbors' are influenced according to Eqn.~\ref{eqn:dnsp}. Therefore the density $\rho^*$ of $x_i$'s neighbors' in $Q$ are recalculated and updated. Thereafter, next sample which holds the lowest dynamic boundary density $\rho^*$ is identified and removed from $Q$. This process continues until $Q$ is empty. At each time of removal, a sequential boundary level is assigned to the samples of being removed. Samples that are removed at the same moment are assigned with the same level. This erosion process is summarized in Alg.~\ref{alg:be}.

\begin{algorithm}
	\KwData{Data sample matrix: $S_{n{\times}d}$}
	\KwResult{Boundary levels: $\Omega$, \textit{r}-NN graph $G$}
	Compute \textit{r}-NN Graph $G$ for $S_{n{\times}d}$\;
	Sort each $G[i]$ in ascending order\;
	Calculate $\rho^*(x_i)$ (Eqn.~\ref{eqn:dnsp}) based on $G$\;
	Push all samples $x_i$s into $Q$\;
	Sort $Q$ in ascending order by $\rho^*$\;
	${\Omega}{\leftarrow}\phi$, $l{\leftarrow}1$\;
	\While{$Q \neq \phi$}{
		Pop $x_i$ with the lowest $\rho^*$ from $Q$\;
		$\Omega{\leftarrow}\Omega{\cup}\{<x_i, l>\}$\;
		\For{each $x_j$ that $x_i$ is in its neighborhood}{
			Calculate $\rho^*(x_j)$\;
			Update $Q$ with $\rho^*(x_j)$\;
		}
		$l{\leftarrow}l+1$\;
	}
	\caption{Produce boundary levels via erosion}
	\label{alg:be}
\end{algorithm}

This erosion process invades inwards from boundaries as more and more samples have been eroded out. It is imaginable that samples that are initially not located on the cluster border are gradually exposed to the boundary erosion. The erosion continues until all the samples have been deleted from $Q$. In the erosion process, a sample living inside automatically ruptures as the start of new boundary when the current lowest dynamic boundary density equals to the density of this sample. An illustration of the erosion process is given by movie \textit{S1} in the supplementary materials.

In the above process, samples are removed out from $Q$ sequentially according to their dynamic boundary density ranging from low to high. Based on the order of being removed from $Q$, sample is assigned with a boundary level $l$, which reflects both the original density $\rho$ (Eqn.~\ref{eqn:dns}) and innerness of one sample as a cluster member. It is easy to see the sample lying outer holds lower boundary level than the one lying inner even they share the same $\rho$. This is the essential difference between our approach and watershed transform~\cite{watershed00}.

\begin{figure}[t]
	\vspace{-0.1in}
	\begin{center}
		\subfigure[density $\rho$]
		{\includegraphics[width=0.42\linewidth]{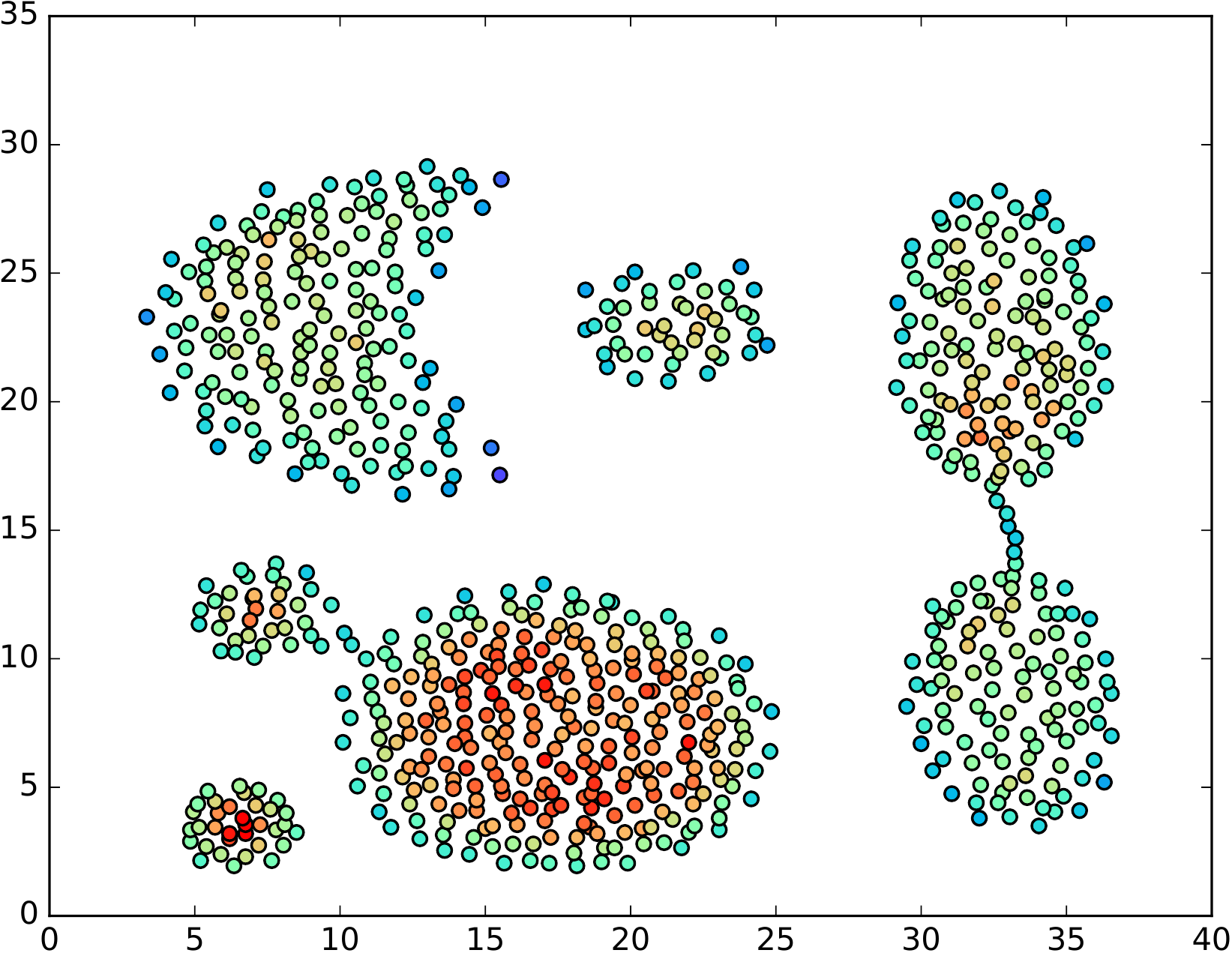}}
		\hspace{0.2in}
		\subfigure[boundary levels]	
		{\includegraphics[width=0.42\linewidth]{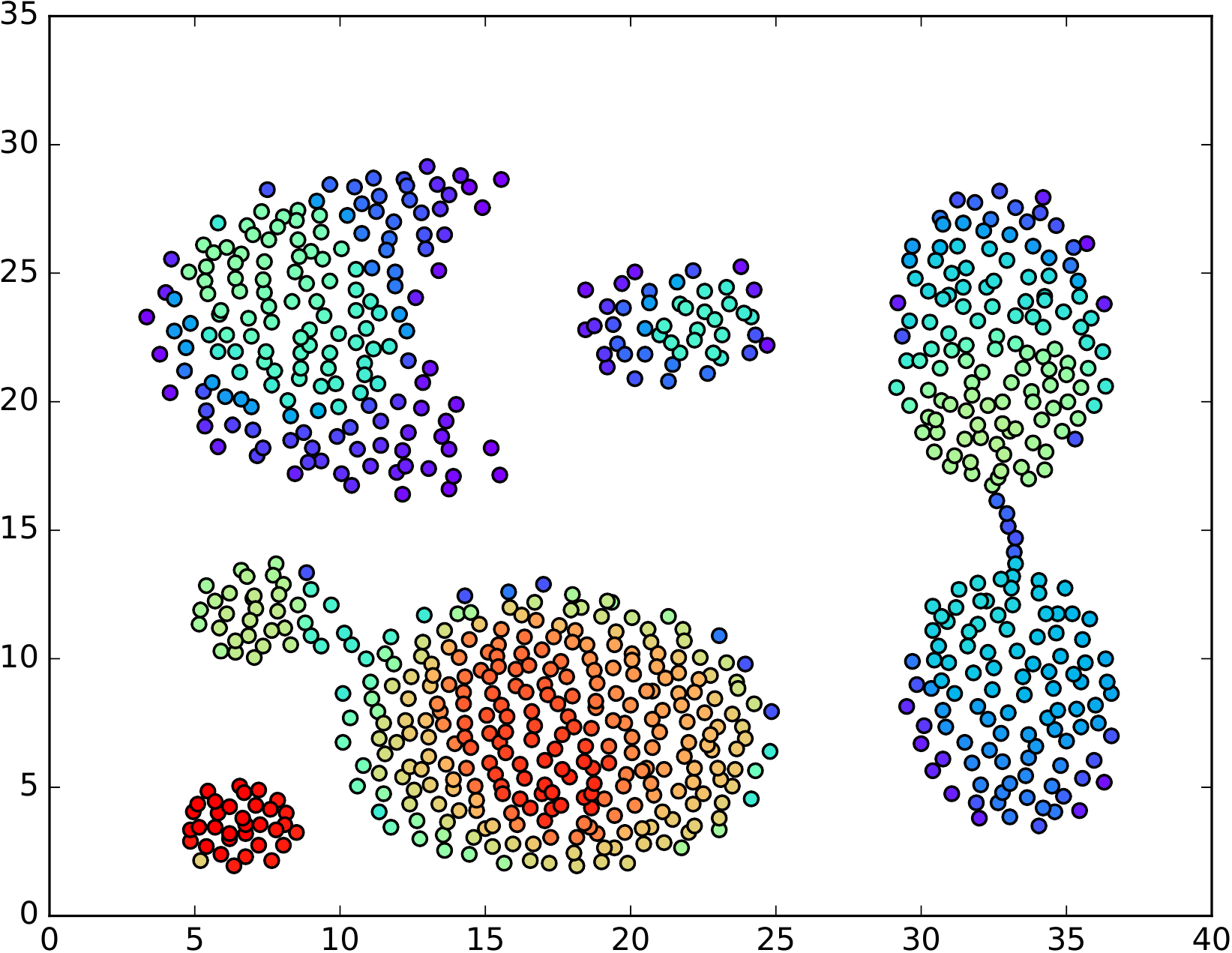}} \\
		\subfigure[3D view of (a)]
		{\includegraphics[width=0.43\linewidth]{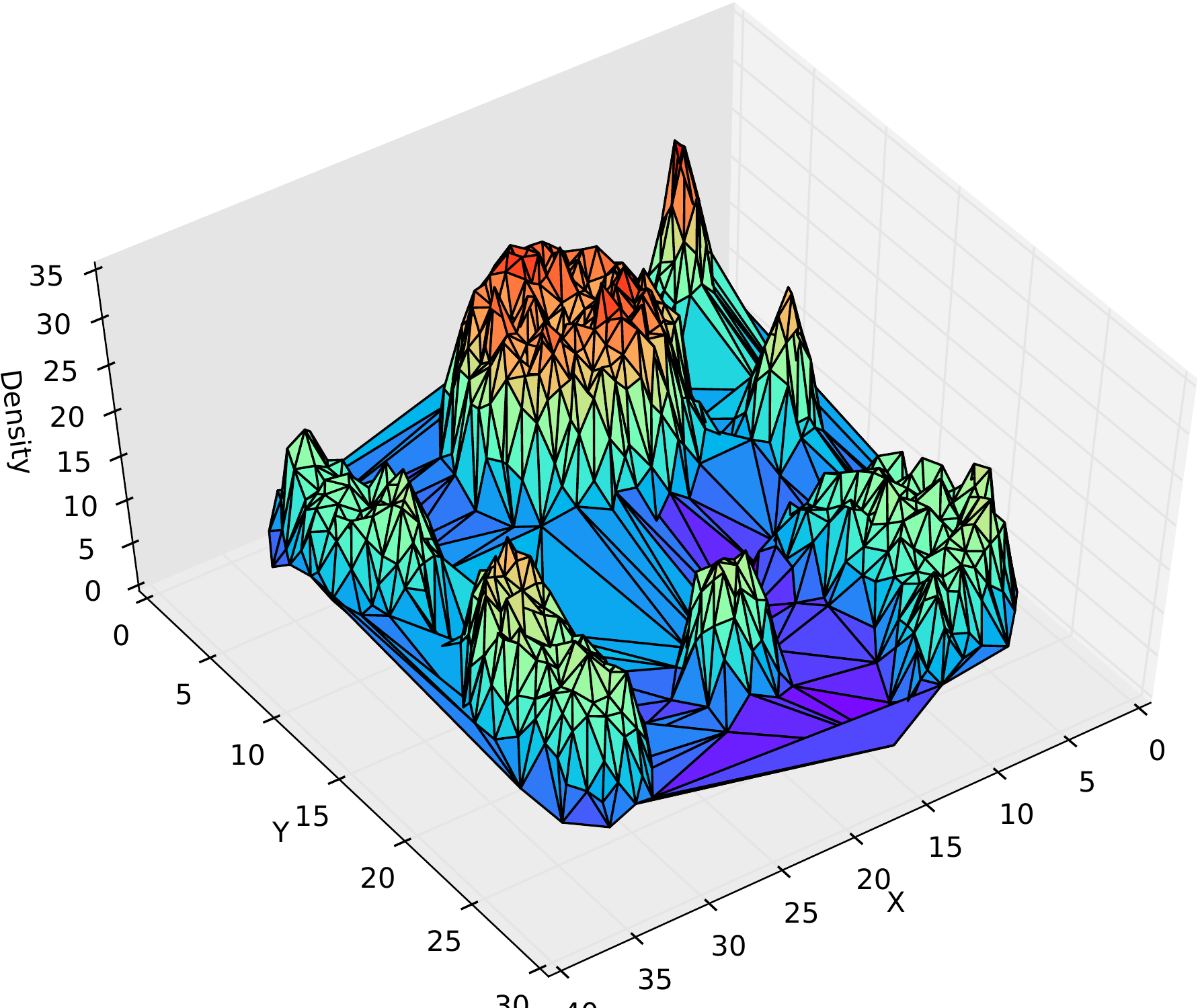}}
		\hspace{0.2in}
		\subfigure[3D view of (b)]
		{\includegraphics[width=0.43\linewidth]{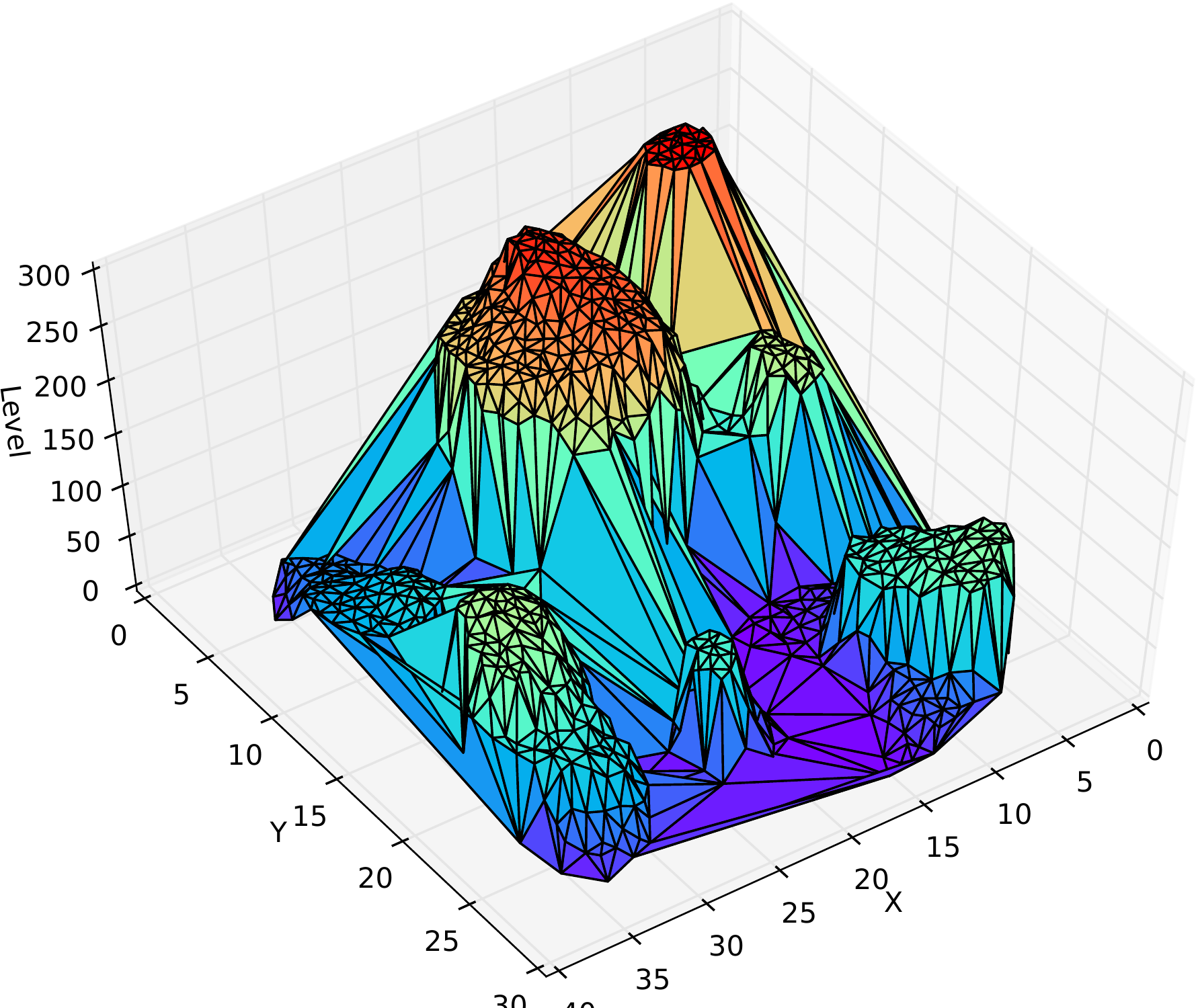}}
		\caption{The comparison between conventional density estimation and sequential boundary levels produced by boundary erosion. The darker of the red color, the higher of the value for both (best viewed in color).
	\label{fig:density}		
		}
	\vspace{-0.15in}
	\end{center}
\end{figure}

Fig.~\ref{fig:density}(a) and Fig.~\ref{fig:density}(b) show the density estimation from Eqn.~\ref{eqn:dns} and the boundary levels produced by the erosion process respectively. Accordingly, the 3D views of Fig.~\ref{fig:density}(a) and Fig.~\ref{fig:density}(b) are shown in Fig.~\ref{fig:density}(c) and Fig.~\ref{fig:density}(d). As shown in the figure, density estimated by Eqn.~\ref{eqn:dns} is full of potholes. This is not surprising since it is not necessarily true that density $\rho$ from the border to the center increases smoothly. The cluster expansion undertaken afterwards is easily trapped in the potholes distributed along the density slope, which is a common issue latent in the traditional approaches. While this issue is avoided in dynamic density estimation, which considers both the density and innerness of one sample. A clear contrast is seen from their 3D views (in Fig.~\ref{fig:density}(c) and Fig.~\ref{fig:density}(d)), the boundary levels produced by Alg.~\ref{alg:be} turn out to be smooth within each emerging cluster. 

In Alg.~\ref{alg:be}, the first step calculates the \textit{r}-NN Graph $G$ which keeps nearest neighbors of each sample within range $r$. $r$ is the only parameter used to set the scale of neighborhood of each sample. Entry $G[i]$ keeps a list of nearest neighbors for sample $x_i$ in its neighborhood $r$. The nearest neighbor list of each entry is sorted in ascending order according to the distances to the sample $x_i$. This will facilitate the afterwards boundary erosion and labeling step (Alg.~\ref{alg:clust}). The time complexity of building nearest neighbor lists for all the samples is quadratic to the scale of input samples, which is on the same level as DBSCAN~\cite{dbscan,dbscan15} and algorithm in~\cite{sci14:alex}. The complexity of computing an approximate \textit{r}-NN graph could be decreased to $O(d{\cdot}n^{1.14})$~\cite{weidong}, which will be discussed in detail in later section.

In order to support fast updating of $\rho^*$ for $x_j$s induced by the removal of $x_i$ (Alg.~\ref{alg:be}, Line 10-13) , a reverse nearest neighbor graph $G^*$~\cite{weidong} is also maintained, in which $G^*[i]$ keeps the data samples $x_j$s that $x_i$ appears in their nearest neighbor lists. Essentially, reverse nearest neighbor graph $G^*$ is nothing more than a simple reorganization of $G$.

\paragraph{Discussion} The advantages of boundary erosion are several folds. Firstly, the erosion takes place on the region of lowest density all the way. It therefore guarantees that the boundaries between clusters are drawn along the most likely regions. From this sense, the global optimality of this process is reached although it is a greedy strategy. Secondly, the bit-by-bit erosion allows the boundaries between clusters to be drawn gradually instead of at once, which is approriate when boundaries are not clear at the beginning. More importantly, the gradual erosion produces an ordered sequence that sorts samples from boundary to center, the reverse of which regularizes a roadmap for cluster expansion. In the whole process, no kernels or heuristic rules are introduced which avoids any unnecessary assumptions on the data distribution or metric spaces.

\subsection{Label Propagation}
Once the sequence of boundary levels are produced, the clustering process becomes natural and could be conveniently undertaken. It is basically a process of cluster expansion that starts from peaks of boundary levels (given in Alg.~\ref{alg:clust}). The propagation starts from the data sample with the highest boundary level. Data sample $x_i$ is assigned with a new cluster label if none of its neighbors in $G[i]$ are labeled. Otherwise, the sample is assigned with the same cluster label as its closest neighbor that has been labeled in the previous rounds. Likewise, the unlabeled samples are sequentially visited following the boundary levels from high to low. The process continues until all the samples are assigned with a label. In this process, the expansion of one cluster stops automatically when it reaches to the cluster boundary, where samples from other cluster hold higher boundary levels. An illustration of this propagation procedure is given by movie \textit{S2} in the supplementary materials.

\begin{algorithm}
	\KwData{Boundary levels: $\Omega$, \textit{r}-NN graph $G$}
	\KwResult{Cluster labels: $L_{n{\times}1}$}
	Sort $\Omega$ by $l$ in descending order\;
	$C{\leftarrow}1$, $L{\leftarrow}0$\;
	\While{$\Omega \neq \phi$}{
		Pop $<x_i, l>$ from $\Omega$\;
		\For{each $x_j$ in $G[i]$}{
			\If{L[j] $>$ 0}
			{
				L[i]${\leftarrow}$L[j];
				break;
			}
		}
		\If{L[i] $==$ 0}
		{ 
			L[i]${\leftarrow}$C\;
			$C{\leftarrow} C + 1$\;
		}
	}
	\caption{Label propagation based on \textit{r}-NN graph}
	\label{alg:clust}
\end{algorithm}

As a summary, the proposed clustering process consists of three steps. Firstly, given the radius of neighborhood $r$, a nearest neighbor graph is built. In the graph, a list of neighbors falling within $r$ range are kept for each sample. With the support of nearest neighbor graph, the sequential boundary levels are produced based on a boundary erosion process in the second step. Finally, clusters are produced by propagating cluster labels sequentially from samples of holding high boundary level to those of lower.

Similar as DBSCAN~\cite{dbscan}, mean-shift~\cite{meanshift} and clusterDP~\cite{sci14:alex}, our algorithm is able to identify clusters of arbitrary shapes as well as the outliers. However, the proposed approach is more attractive from several aspects of view. On one hand, unlike DBSCAN, no heuristic rules are introduced, which makes the clustering insensitive to extra the parameter settings. On the other hand, unlike mean-shift or clusterDP in~\cite{sci14:alex}, no kernel is adopted in the density estimation, which makes it feasible for various types of metric spaces. Moreover, unlike DBSCAN or clusterDP in~\cite{sci14:alex}, no cluster centers or cluster peaks are explicitly defined or specified. Instead, similar as affinity propagation~\cite{sci07:frey}, the cluster peaks and the clusters emerge gradually. Furthermore, in the algorithm, there is no specification on the distance measure. As a consequence, unlike \textit{k}-means~\cite{km82}, mean-shift~\cite{meanshift} or recent OCTD~\cite{ocdt16:zdyu}, it is feasible for various metric spaces as long as the density of samples could be estimated.

The boundary erosion shares similar motivation as ``border-peeling'' in~\cite{peel17:nadav}, however they are essentially different in three major aspects. Firstly, no kernel is introduced in our approach. Secondly, all samples will be eroded out after the erosion process. In constrast, cores points are reserved for cluster expansion in ``border-peeling''. Finally, ``border-peeling'' relies on DBSCAN to reconstruct the clusters, while clustering in boundary erosion is undertaken via label propagation with the guidance of a \textit{r}-NN graph.

In the above label propagation process, the same \textit{r}-NN graph is used as the boundary erosion process (Alg.~\ref{alg:be}). Alternatively, it is feasible to use a different \textit{r}-NN graph in the label propagation. In some cases, the density of a sample is very low. Such kind of samples are usually recorgnized as outliers by Alg.~\ref{alg:be}. However in certain scenario, we may expect that such kind of outliers are assigned to clusters that are the most close to them. To achieve that, the \textit{r}-NN graph that is supplied to the above expansion procedure is revised. In particular, $G[i]$ is augmented to top-$k$ nearest neighbors when the size of its nearest neighbors list is less than $k$, where $k$ is another given parameter. In the experiment section, we are going to show this \textit{augmented propagation} strategy is meaningful in certain circumstances.

According to our observation, boundary erosion fails only when samples from different clusters are mixed with each other. In this case, the assumption of this algorithm that clusters are separated by sparse regions actually breaks. However, it is possible to address this issue by recent sub-space embedding~\cite{nips17:deepssc}. The input high dimensional data are firstly projected to lower and separable space by DSC-Net-L2~\cite{nips17:deepssc}. Boundary erosion is therefore applied on the projected data, which will be illustrated in the experiment section.

\section{Clustering in Large-scale}
\label{sec:lgclust}
As presented in Alg.~\ref{alg:be}, \textit{r}-NN graph is required as the pre-requisite of the boundary erosion process. The time complexity of calculating \textit{r}-NN graph could be as high as $O(d{\cdot}n^2)$. Moreover, in the worst case, the space complexity of keeping \textit{r}-NN graph is close to $O(n^2)$ since one could not assume how many neighbors are located in range $r$ in advance. As a consequence, this algorithm becomes computationally inefficient in the situation that both $n$ and $d$ are large. To address this issue, an approximate solution is presented in this section.

\subsection{Clustering with Approximate \textit{r}-NN Graph}
\label{sec:apprnn}
As shown above, it is computationally expensive to calculate an exact \textit{r}-NN graph particularly in high dimensional and large-scale cases. Many attempts have been made to seek for approximate solutions for this issue. Thanks to the progress made in recent years, with NN-Descent algorithm presented in~\cite{weidong}, it is possible to construct a \textit{k}-NN graph in high accuracy under the empirical complexity of $O(d{\cdot}n^{1.14})$. More attractively, there is no specification on distance measure in the algorithm, which is precisely in line with our clustering algorithm.

In our practice for large-scale clustering task, the first step of Alg.~1 (i.e., Line 1) is modified.  NN-Descent~\cite{weidong} is called to produce an approximate \textit{k}-NN graph. The \textit{k}-NN list of each sample is further pruned according to the given parameter $r$, which results in an approximate \textit{r}-NN graph . While the rest of clustering process remains unaltered. In the experiment section, the results on the large-scale image clustering are illustrated.

\subsection{Complexity Analysis}
\label{sec:cmp}
As presented in the previous sections, clustering via boundary erosion basically consists of three steps. In the first step, an \textit{r}-NN graph is built. The time complexity of building an exact \textit{r}-NN is $O(d{\cdot}n^2)$. This is feasible for low dimensional and small scale cases. While for high dimensional and large-scale task, NN-Descent~\cite{weidong} is adopted for the approximate \textit{r}-NN graph construction, for which the construction complexity is around $O(d{\cdot}n^{1.14})$~\cite{weidong}. In the second step, the boundary erosion process operates on a dynamic array $Q$. Each time, at least one sample with the lowest boundary density is removed from the array. $\bar{k}$ samples on average\footnote{This is the average number of samples fall into neighborhood $r$.} in the array, that are influenced by the removal of one sample, are updated. For efficiency, the dynamic array can be implemented with a heap. The removal repeats for at most $n$ times. As a result, the time complexity of this step is $\sum_{i=0}^{n}log(n+i{\cdot}{\bar{k}})$, which is on $O(n{\cdot}{log(n)})$ level. In the label propagation step, it is clear to see the complexity is only $O(n)$. Overall, the complexity of the clustering algorithm is $O(d{\cdot}n^2+n{\cdot}log(n))$ if one expects exact solution. This is suitable for small scale tasks. While for large-scale cases, the complexity is only $O(d{\cdot}n^{1.14}+n{\cdot}log(n))$ with the support of NN-Descent \textit{k}-NN graph construction, which is even more efficient than the conventional \textit{k}-means.

\section{Experiments}
\label{sec:exp}
In the following, the performance of the proposed boundary erosion (BE) is studied in comparison to several state-of-the-art approaches on various evaluation benchmarks and tasks, such as synthetic data of different distributions, face image grouping, and clustering on biological as well as large-scale image data. On large scale image clustering part, our algorithm is implemented with C++ and compiled with \textit{GCC 5.4},and conducted by single thread on a PC with \textit{3.6GHz} CPU and \textit{16G} memory setup.
\subsection{Clustering on Synthetic Datasets}
\begin{figure}[t]
	\begin{center}
	\vspace{-0.1in}
		\subfigure[]
		{\includegraphics[width=0.41\linewidth]{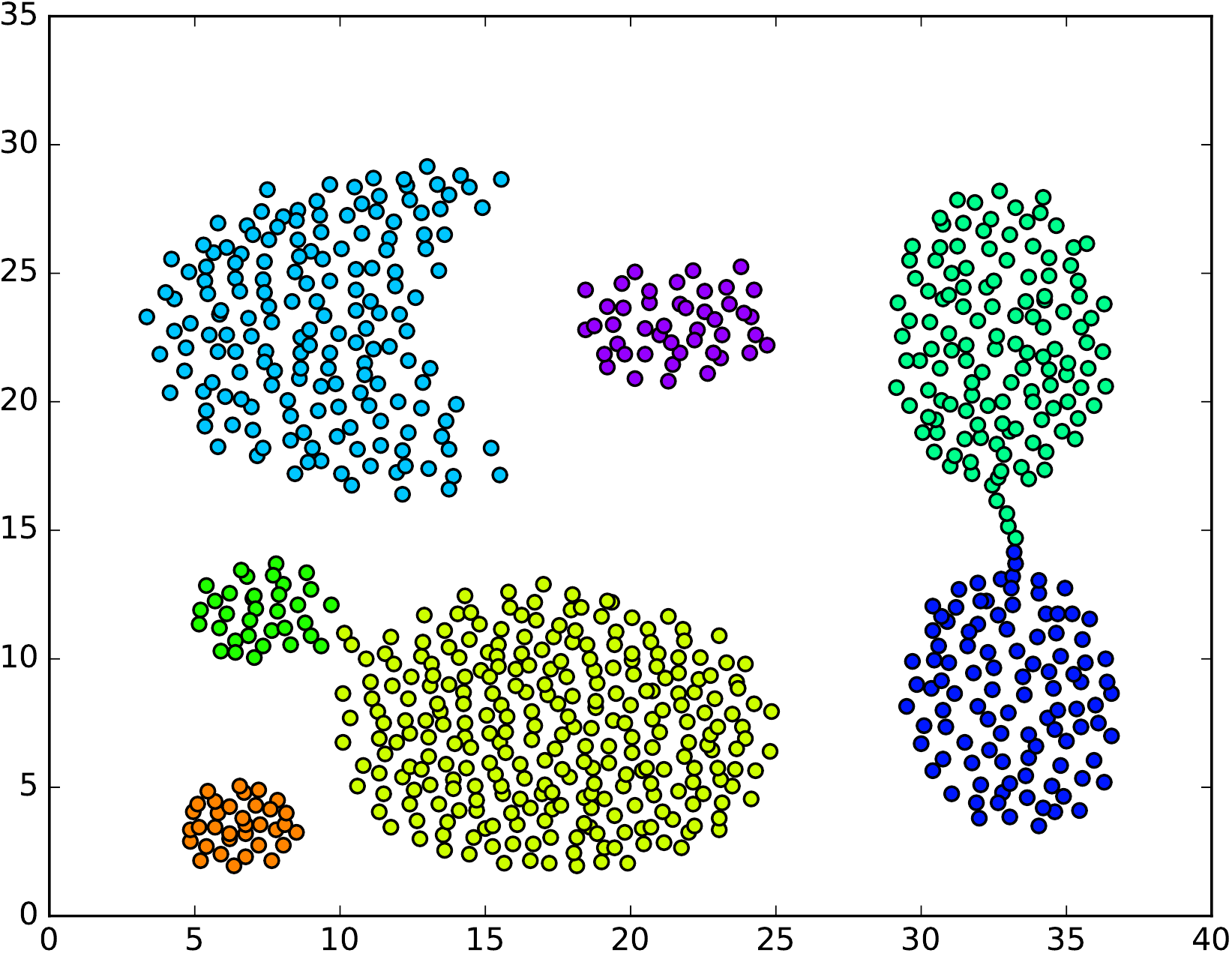}}
		\hspace{0.1in}
		\subfigure[]
		{\includegraphics[width=0.45\linewidth]{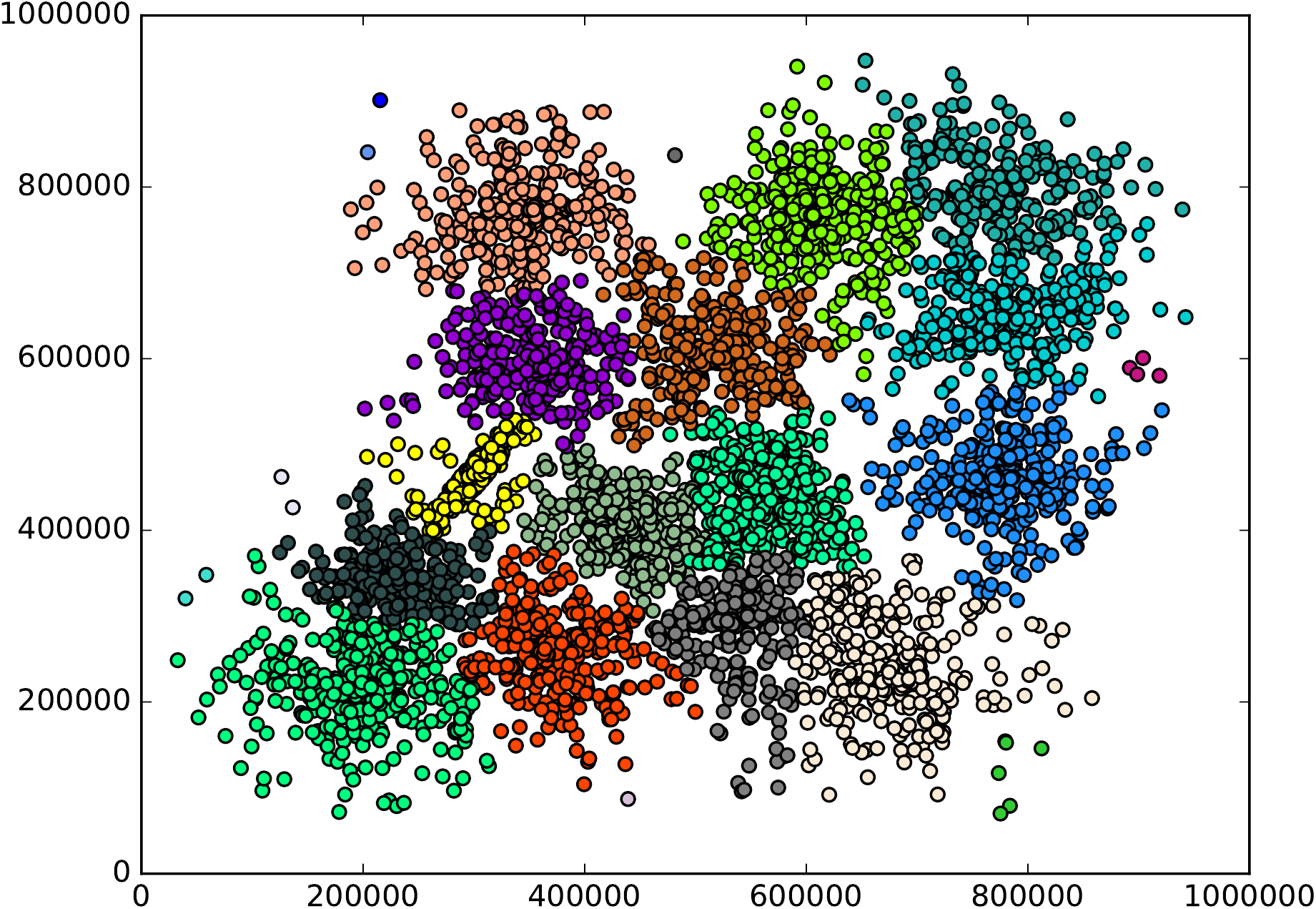}}\\
		\hspace{-0.15in}
		\subfigure[]
		{\includegraphics[width=0.41\linewidth]{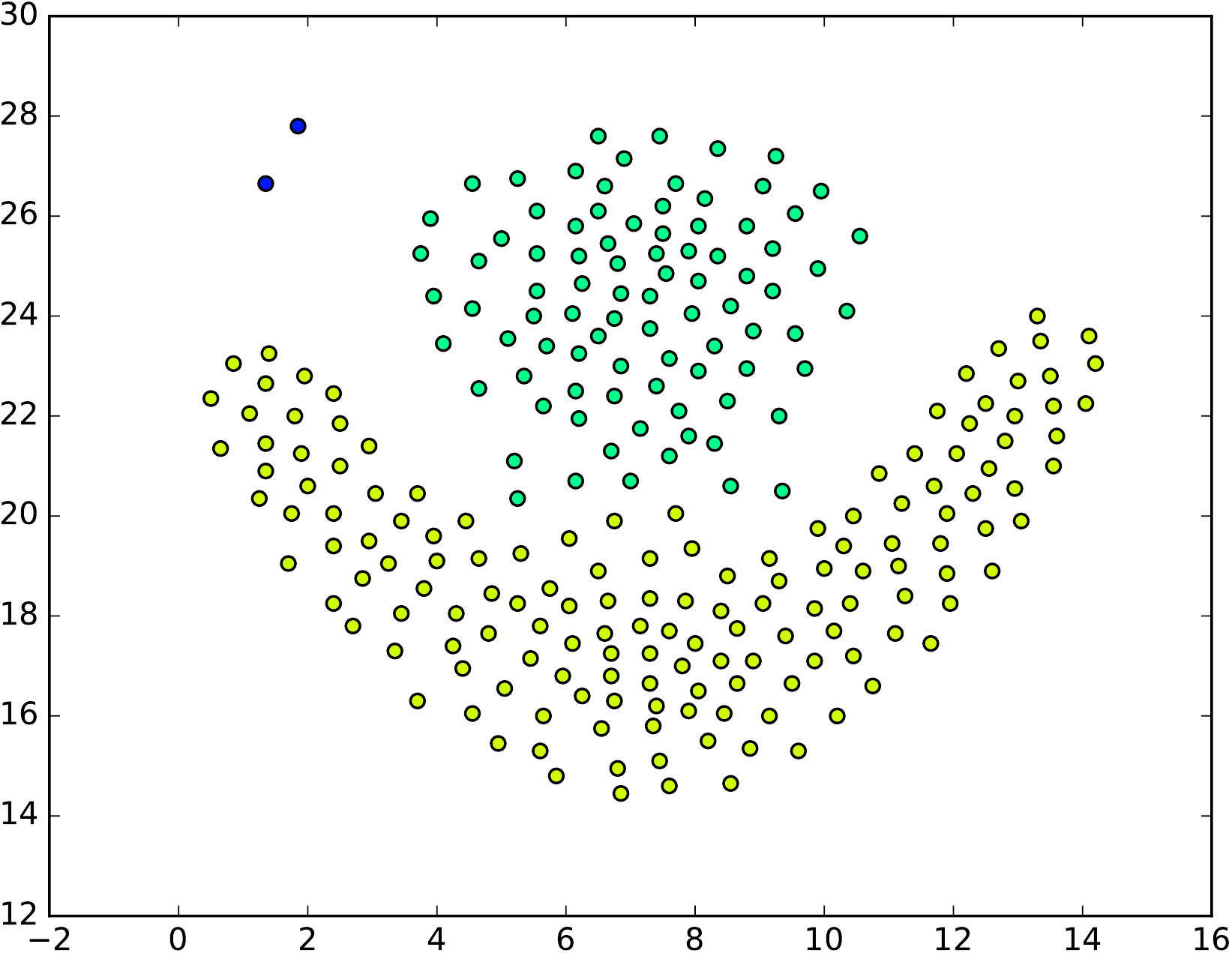}}
		\hspace{0.15in}
		\subfigure[]
		{\includegraphics[width=0.41\linewidth]{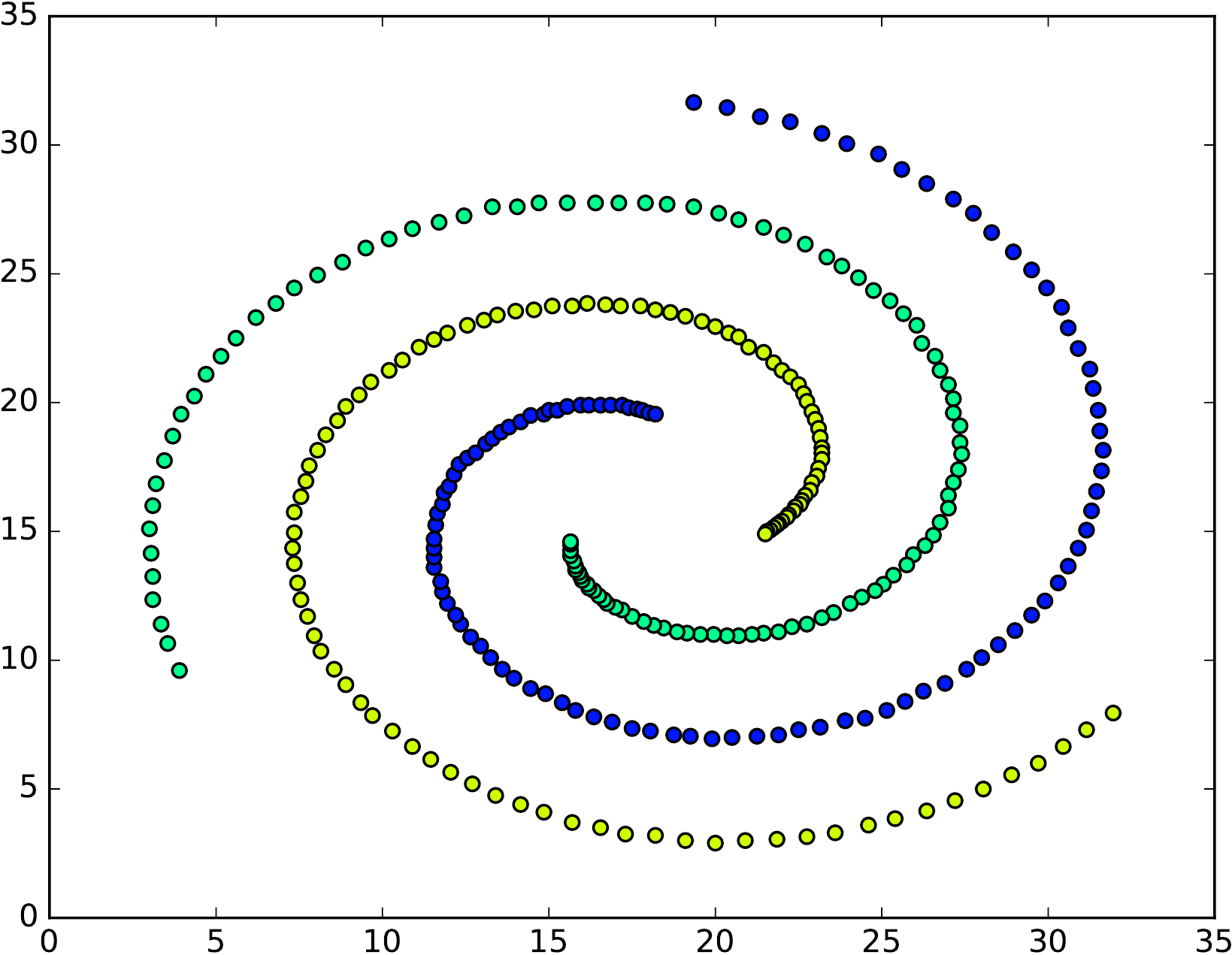}}\\
		\subfigure[]
		{\includegraphics[width=0.41\linewidth]{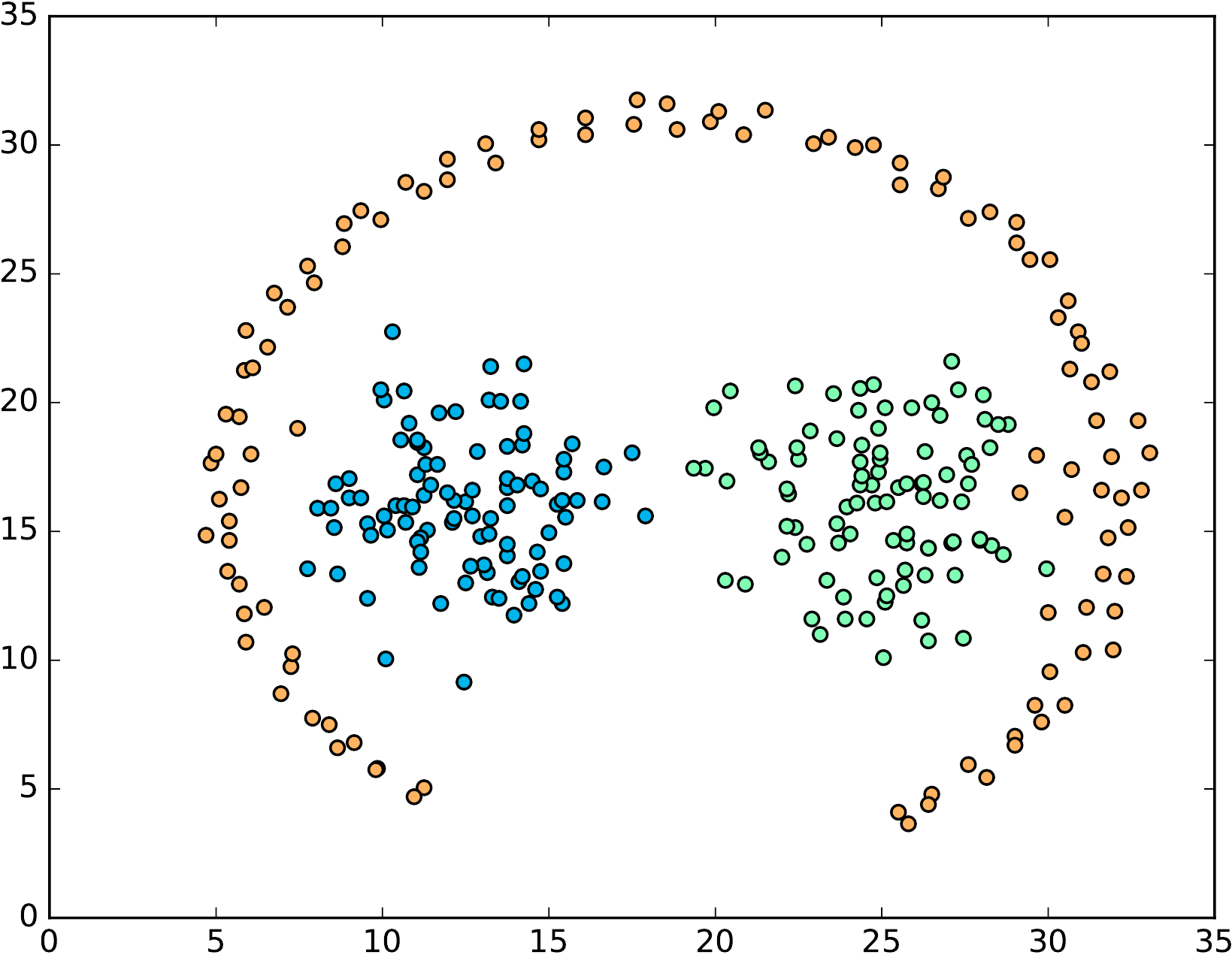}}
		\hspace{0.15in}
		\subfigure[]
		{\includegraphics[width=0.41\linewidth]{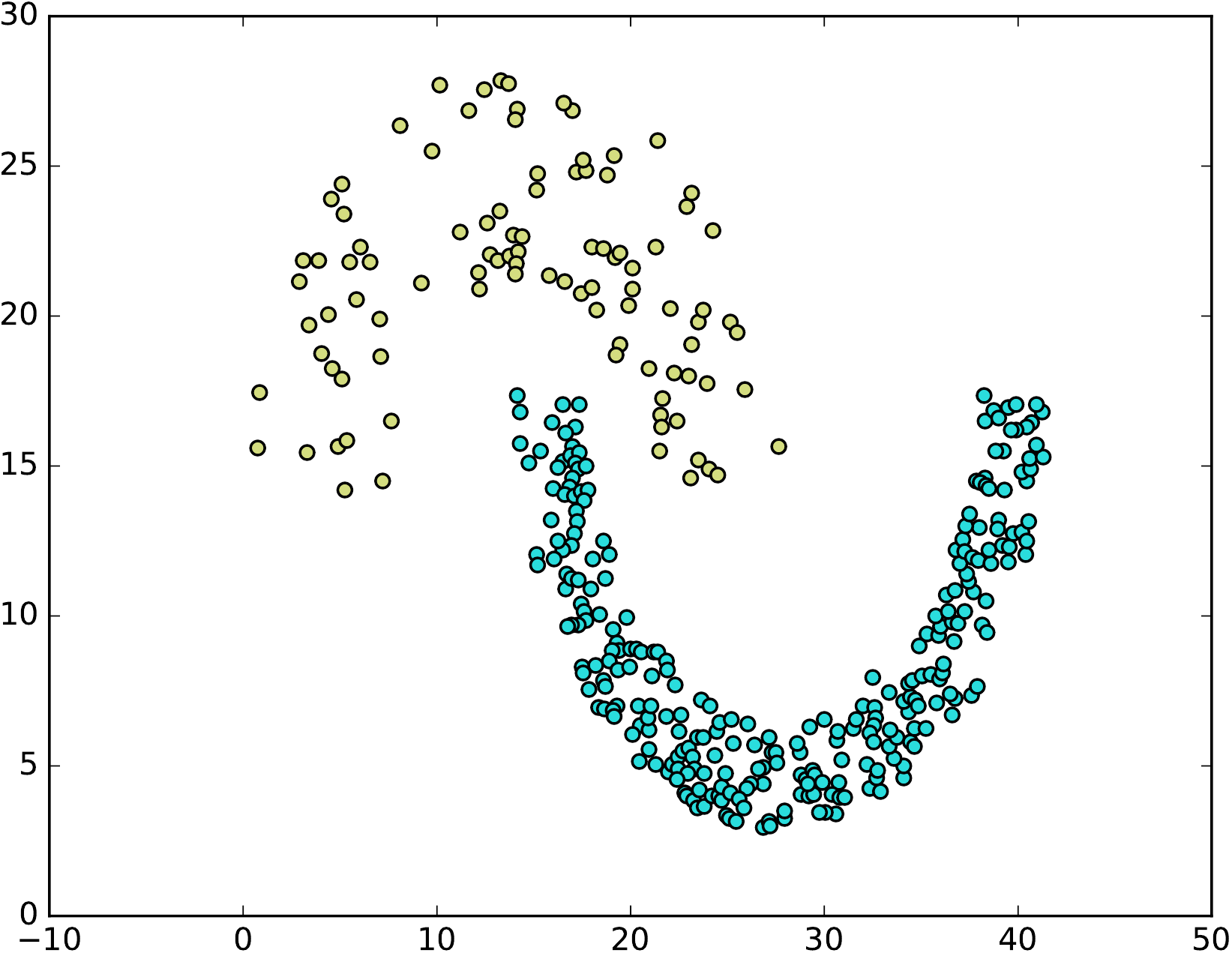}}
		\caption{Clustering results on six synthetic datasets (best viewed in color). The datasets in Fig~\ref{fig:synt}(a to f) are namely Aggregation (AGG)~\cite{kdd07:gionis}, S3~\cite{pr06:franti}, flame~\cite{bmc07:fu}, sparil \cite{pr08:chang}, Jain~\cite{jain} and pathbased (Path)~\cite{pathbased} respectively. The valid range of $r$ that allows to reproduce the same results as shown are [\textit{1.5}, \textit{2.4}], [\textit{3.5}${\times}{10^4}$, $\textit{4.8}{\times}10^4$], [\textit{1.3}, \textit{2.6}], \textit{2.5} and \textit{3.8} for \textit{a},  \textit{b}, \textit{c}, \textit{d}, \textit{e}, and \textit{f} respectively.
	\label{fig:synt}		
		}
	\vskip -0.25in
	\end{center}
\end{figure}

\begin{figure}[t]
	\vspace{-0.1in}
	\begin{center}
		\subfigure[]
		{\includegraphics[width=0.41\linewidth]{./data2.pdf}}
		\hspace{0.1in}
		\subfigure[]
		{\includegraphics[width=0.45\linewidth]{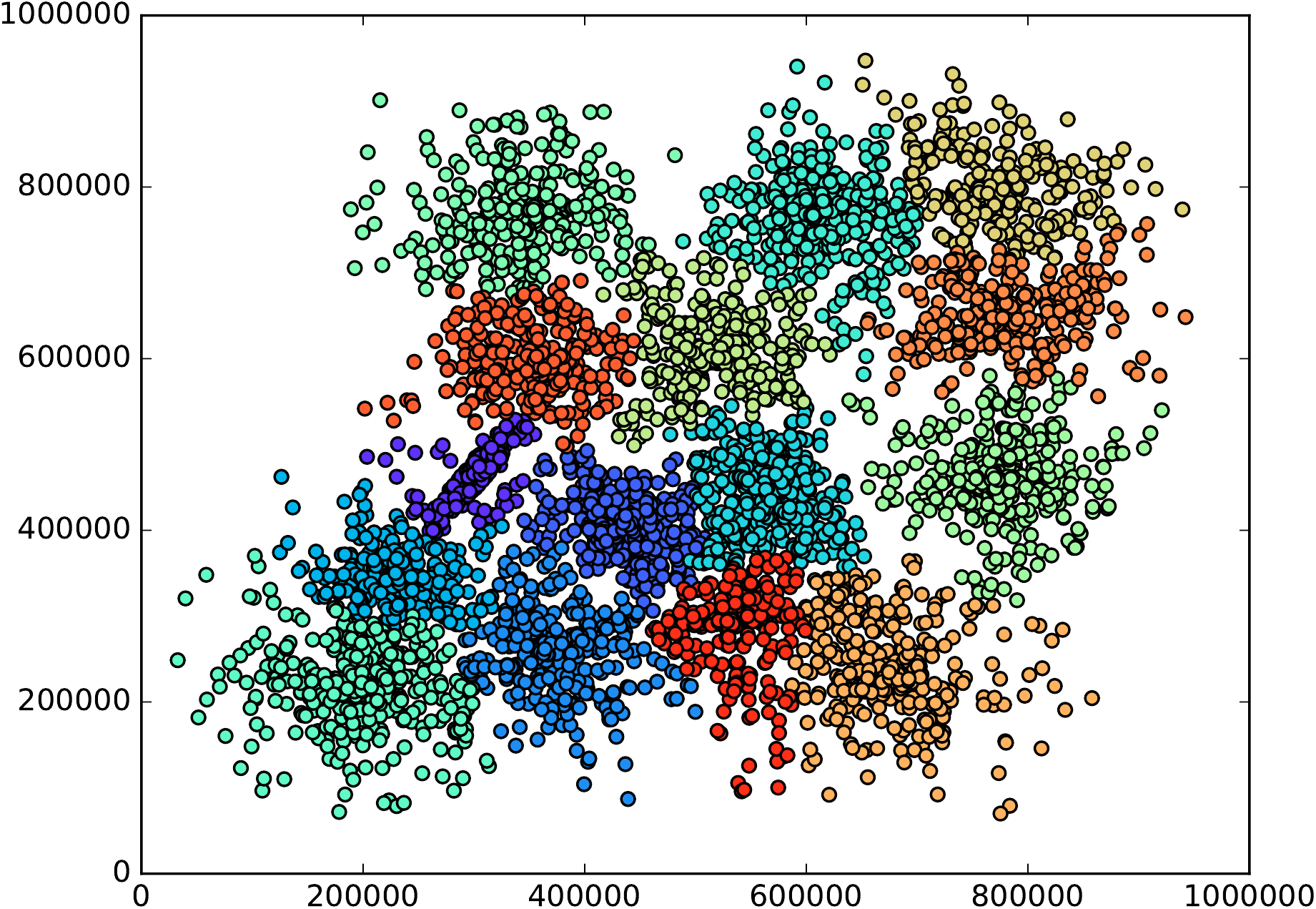}}\\
		\hspace{-0.15in}
		\subfigure[]
		{\includegraphics[width=0.41\linewidth]{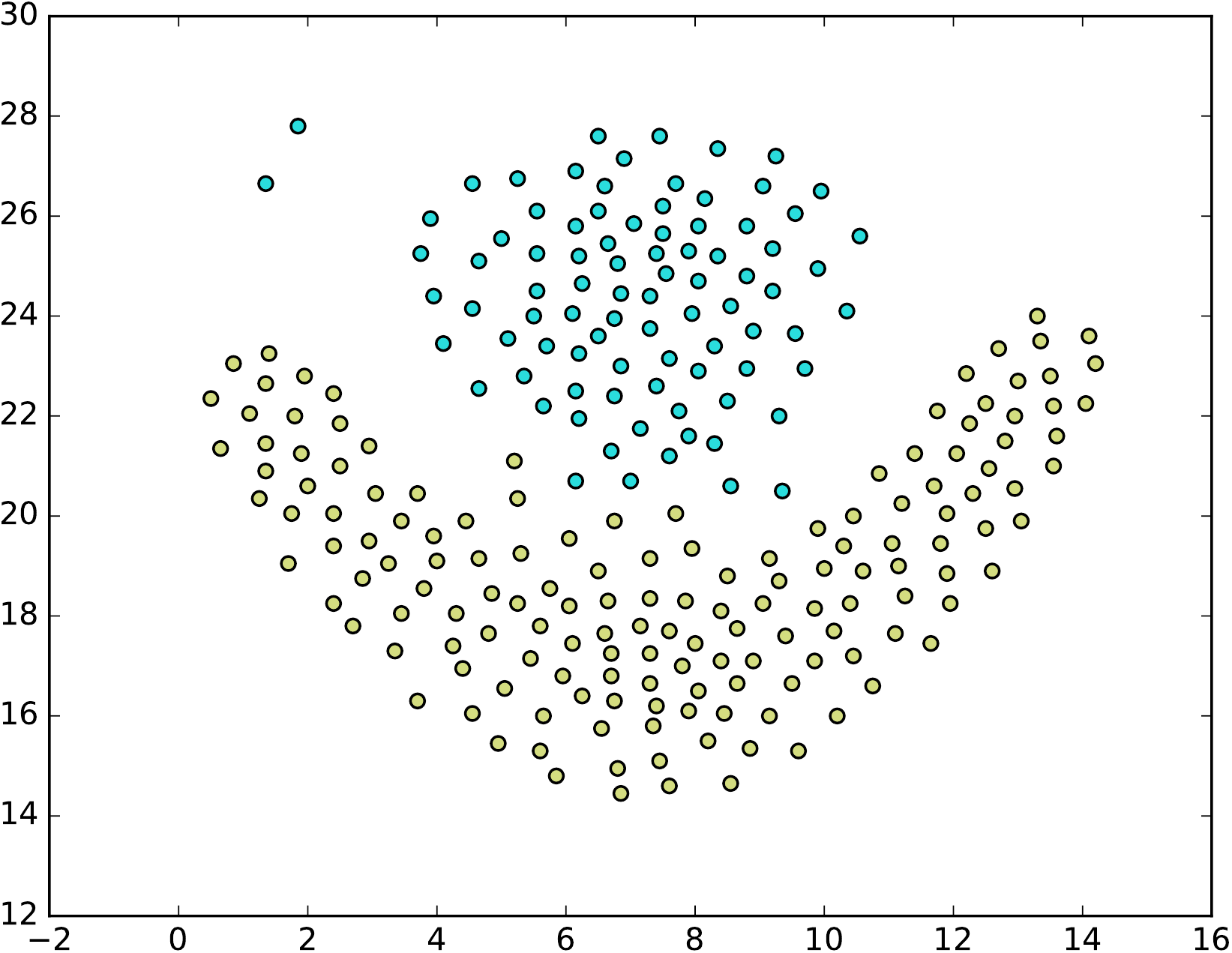}}
		\hspace{0.15in}
		\subfigure[]
		{\includegraphics[width=0.41\linewidth]{./sparil.pdf}}\\
		\subfigure[]
		{\includegraphics[width=0.41\linewidth]{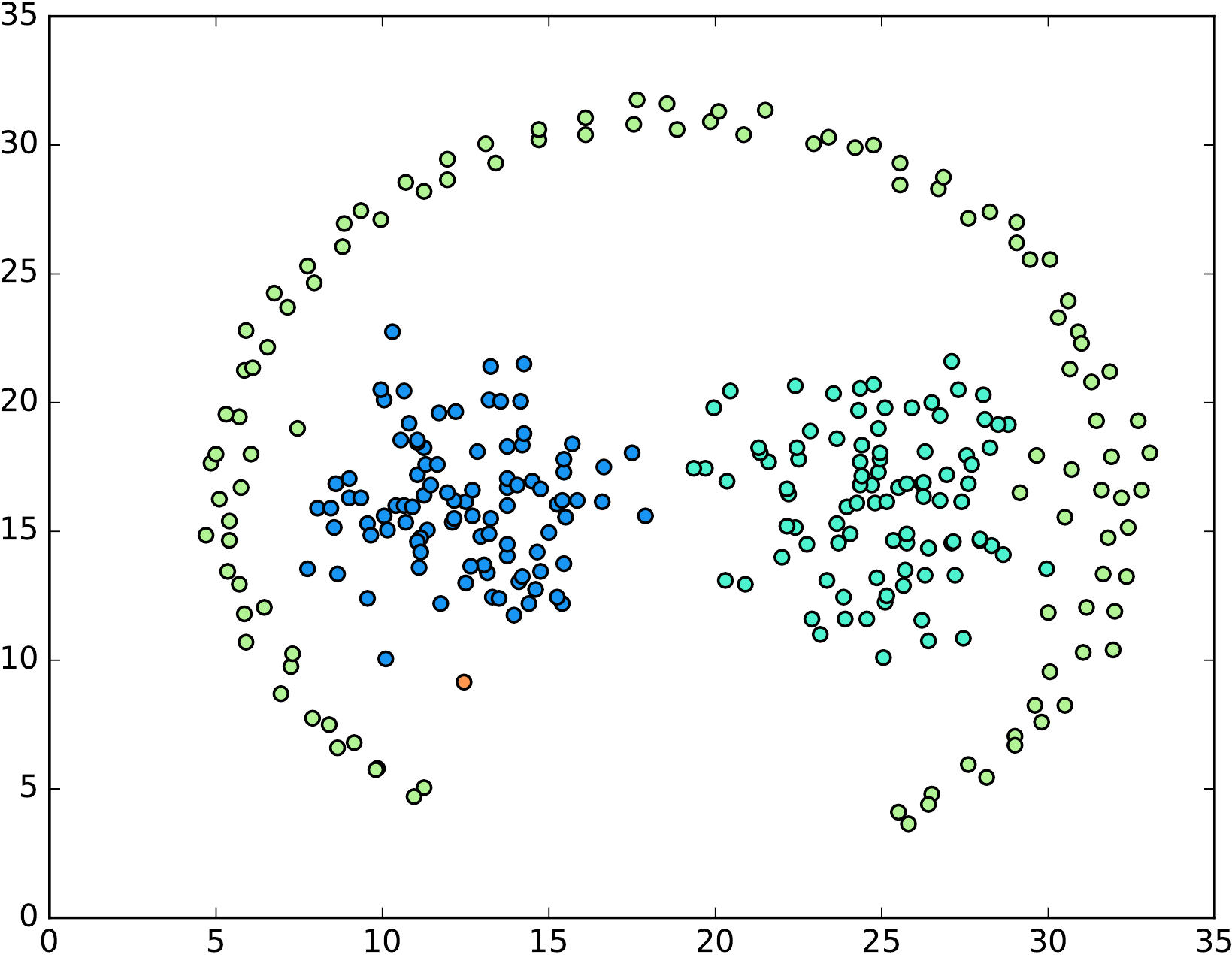}}
		\hspace{0.15in}
		\subfigure[]
		{\includegraphics[width=0.41\linewidth]{./jain.pdf}}
		\caption{Clustering results on six synthetic datasets (best viewed in color) by boundary erosion with \textit{augmented propagation}. The valid range of $r$ that allows to reproduce the same results as shown are [\textit{1.4}, \textit{2.4}], [\textit{2.6}${\times}{10^4}$, $\textit{4.8}{\times}10^4$], [\textit{1.0}, \textit{2.7}], [\textit{1.3}, \textit{4.1}],[\textit{2.1}, \textit{2.5}] and \textit{3.8} for \textit{a},  \textit{b}, \textit{c}, \textit{d}, \textit{e}, and \textit{f} respectively.
	\label{fig:asynt}}
		\vskip -0.25in
	\end{center}
\end{figure}

The first experiment is conducted on six synthetic datasets, all of which are 2D spatial data points. The data points are drawn from a probability distribution with nonspherical shapes. These datasets have been widely adopted to test the robustness of a clustering algorithm. Results produced by our algorithm are presented in Fig.~\ref{fig:synt}. The valid range of parameter $r$ for each dataset that allows to reproduce the same results is accordingly attached. As shown in the figure, the proposed algorithm is able to identify all the clusters as well as the outliers of each case. In particular, satisfactory results are observed on challenging datasets \textit{S3}, \textit{Path} and \textit{Jain}, on which existing approaches hardly produce descent results. 

Fig.~\ref{fig:asynt} shows the results from the \textit{augmented propagation} strategy. As shown from the figure, the results for datasets \textit{a}, \textit{d} and \textit{f} are the same as previous experiment. While for datasets \textit{b}, \textit{c} and \textit{e}, where the outliers are in presence, the \textit{augmented propagation} assigns the outliers to the clusters that are the most close to them. This is meaningful in the case that one prefers to producing clusters without isolated outliers. The results in Fig.~\ref{fig:asynt}  are also quantiatively shown in Tab.~\ref{syn_table} in terms of clustering accuracy~\cite{ocdt16:zdyu}. It is compared to \textit{k}-means (KMS), spectural clustering (SC) and order-constrained transitive distance clustering (OCTD). Perfect results are achieved on most of the datasets. Compared to results from the most representative methods (in~\cite{ocdt16:zdyu}), BE achieves the best performance in all the cases.

In the following experiments, the results are produced by \textit{r}-NN graph without augmentation if it is not specified. 

\begin{table}
\caption{Clustering accuracy (\%) on synthetic datasets. For BE, the augmented \textit{r}-NN graph is adopted in label propagation, which allows isolated samples to be assigned to its closest cluster}
\label{syn_table}
	\vskip 0.1in
\begin{center}
\scriptsize{
\begin{sc}
\begin{tabular}{l|ccccccr}
\hline
Mthd. & Agg  & S3 & flame  & sparil & Path & Jain  \\
\hline\hline
Kms&87.92& 85.58 &84.17&33.97&74.34&78.28\\
SC&99.37&8.10 &98.75&59.30&97.00&\textbf{100.00}\\
OCTD&99.87&U.A. &\textbf{100.00}&\textbf{100.00}&96.66&\textbf{100.00}\\
\hline
BE   & \textbf{100.00} & \textbf{95.80}  & \textbf{100.00} & \textbf{100.00} & \textbf{100.00}  &\textbf{100.00}&  \\
\hline
\end{tabular}
\end{sc}
}
\end{center}
\vskip -0.1in
\end{table}

\subsection{Clustering on Biological Data}
\begin{table}
	\centering
	\caption{Comparisons to state-of-the-art algorithms on Brown dataset. Results of the state-of-the-art algorithms are cited from~\cite{nature15}}
	\vskip 0.1in
	\scriptsize{
	\begin{sc}
	\begin{tabular}{lcl}\hline
		Mthd. & $F_1$-score & Para. Settings\\\hline \hline
		DIANA&	0.991&	\textit{metric = $l_2$}, \textit{k = 26}\\
		AGNES&	0.987&	\textit{Complete-link},\\&& \textit{metric = $l_2$},\textit{k = 25}
		\\
		HC&	0.987 &	\textit{Complete-link},\\&& \textit{k = 25}
		\\
		TC	&0.986&	\textit{T = 48.868}
		\\
		clusterDP&	0.975&	\textit{k = 25},\\&& \textit{dc = 258.645}
		\\
		clusterONE&	0.946&	\textit{s = 1}, \textit{d = 0.0}\\
		%		MCODE&	0.931&	\textit{v=0, cutoff=40.413},\\&& \textit{haircut=T,fluff=T}\\\hline
		MC&	0.923&	\textit{I = 2.196}\\
		\textit{k}-Medoids&	0.912&	\textit{k = 37}\\
		AP&	0.910&	\textit{dampfact = 0.845},\\&&\textit{preference = 80.827}\\&&	\textit{maxits = 5000},\\&& \textit{convits = 500}\\
		DBSCAN&	0.680&	\textit{eps = 323.306},\\&& \textit{MinPts = 1}\\
		SC&	0.656 &	\textit{k = 11}\\\hline \hline
		BE &	\textbf{0.998} & \textit{r = 60}	\\\hline
	\end{tabular}
	\end{sc}
	}
	\label{tab:brown}
\end{table}
In this part, our algorithm is tested on Brown dataset~\cite{nature15} which are biological data. In this datset, the affinity matrix that keeps pairwise distances between DNA sequences are supplied. They are pairwise of blasted sequences of \textit{232} proteins belonging to \textit{29} groups of families. In this case, algorithms such as \textit{k}-means, are not feasible since they only work on $l_2$-space. In this study, BE is compared to DIANA~\cite{agnes90}, AGNES~\cite{agnes90}, Hierarchical Clustering (HC)~\cite{hierarchical}, Transitivity Clustering (TC)~\cite{transitive}, clusterDP~\cite{sci14:alex}, clusterONE~\cite{clusterone}, Markov Clustering (MC)~\cite{markov}, \textit{k}-Medoids (PAM)~\cite{agnes90}, Affinity Propagation (AP)~\cite{sci07:frey}, DBSCAN~\cite{dbscan} and Spectral Clustering (SC)~\cite{normcut}.

\textit{Twenty eight} clusters are produced when $r = 60$ by our algorithm. The $F_1$ score is \textit{0.998}, which is nearly perfect. This is also the best performance ever reported according to~\cite{nature15}, as is shown in Table~\ref{tab:brown}. While affinity propagation~\cite{sci07:frey} only achieves \textit{0.910}, which is considerably worse than that of our algorithm. Our algorithm also outperforms clusterDP~\cite{sci14:alex} by more than \textit{2\%}.

\subsection{Clustering on High-dimensional Data}
\begin{figure}
\centering{
		\includegraphics[width=0.90\linewidth]{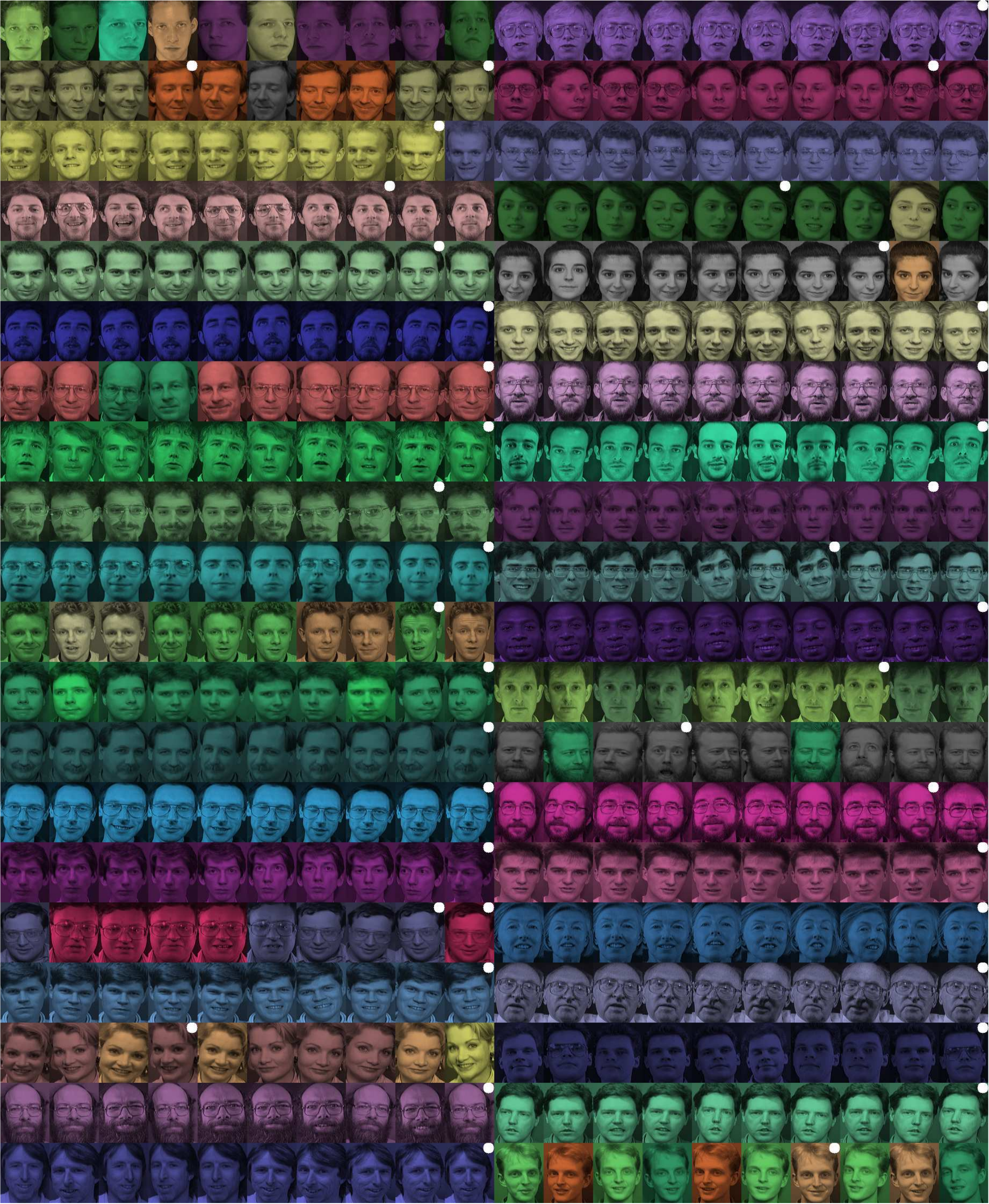}}
		\caption{Clustering results on the \textit{40} groups of Olivetti Face Database (best viewed in color). Faces with the same cover color are clustered into the same cluster. The cluster centers are labeled with a white circle. 
	\label{fig:face}}
\end{figure}

Our algorithm is also tested on two face datasets namely Olivetti Face Database (ORL)~\cite{face94} and Extended Yale B(EYaleB)~\cite{eyaleb}, and two visual object image datasets, namely COIL20~\cite{coil20} and COIL100~\cite{coil100}. In these four datasets, there are \textit{400}, \textit{2432}, \textit{1440} and \textit{7400} images which are from \textit{40}, \textit{38}, \textit{20} and \textit{100} visual object groups respectively. On these four datasets, clustering algorithms are expected to identify images that are from the same object groups. Since the images are not directly separable by their pixel intensities (i.e., RGB), images are projected to low-dimensional feature space by DSC-Net-L2~\cite{nips17:deepssc}. Our algorithm (BE) is adopted in the final clustering stage. In the experiments, DSC-Net-L2 in combination with BE (denoted as DSC+BE) is compared to sparse spectral clustering (SSC)~\cite{ssc}, DSC-Net-L2 in combination with spectural clustering (denoted as DSC+SC) and standard configuration of DSC-Net-L2 based clustering~\cite{nips17:deepssc}, in which a discriminative variant of spectral clustering is integrated. The clustering error rates~\cite{nips17:deepssc} are shown in Table~\ref{tab:orl}. As shown in the table, DSC+BE outperforms or is very close to the best results ever reported on these datasets.

 \begin{table}
 \caption{Clustering error rates on four image datasets. The radius $r$ of BE is set to \textit{0.55}, \textit{0.55}, \textit{0.06} and \textit{0.06} respectively on four datasets}
 \label{tab:orl}
	\vskip 0.1in
 \begin{center}
 \scriptsize{
 \begin{sc}
 \begin{tabular}{lcccccr}
 \hline
 Datasets &  SSC  & DSC&DSC+SC & DSC+BE  \\
 \hline
 ORL     &  32.50  & 14.00   &15.16& \textbf{12.23}    \\
 EYaleB  &  27.51 & \textbf{2.67} &11.92& 4.52    \\
 COIL20  &  14.86 & 5.14 &9.00 & \textbf{3.82}    \\
 COIL100 &  45.00 & \textbf{30.96}&34.99& 31.67    \\
 \hline
 \end{tabular}
 \end{sc}
 }
 \end{center}
 \vskip -0.1in
 \end{table}
 
Overall, superior performance is achieved in all experiments and on different categories of data by BE, which is essentially attributed to its extraordinary capability of identifying clusters in arbitrary shapes and the genericness of its model.

\subsection{Image Clustering in Large-scale}
In this section, the effectiveness of the proposed clustering algorithm is verified on image clustering/linking task. A subset of YFCC100M~\cite{yfcc} is adopted for evaluation. There are \textit{1.1} million images in total. They are represented with deep features from HybridNet~\cite{mm16:amato}, which are in \textit{128} dimensions after PCA. In the clustering, NN-Descent is called to build the approximate \textit{r}-NN graph for YFCC \textit{1.1} million. In the experiments, top-$k$ is fixed to \textit{5} for  YFCC \textit{1.1} million. While $r$ is set to \textit{0.70}. The \textit{augmented propagation} is adopted in the cluster expansion stage, which avoids isolating similar images that are under severe transfomations.

\begin{figure}
	\begin{center}	
		{\includegraphics[width=0.9\linewidth]{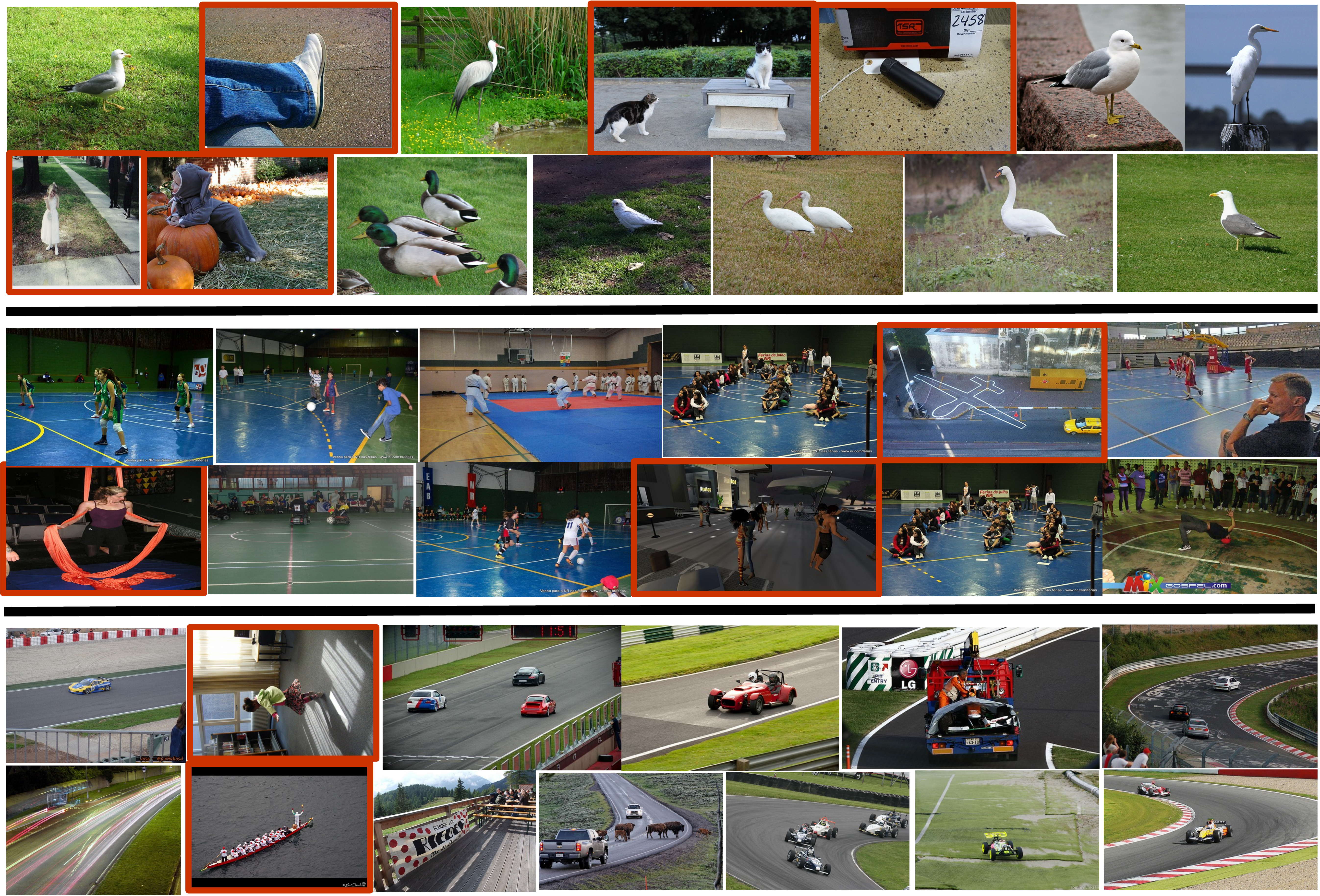}}
		\caption{Sample image groups that have been successfully identified by our algorithm on YFCC \textit{1.1} million).
	\label{fig:paris}		
		}
	\vspace{-0.2in}
	\end{center}
\end{figure}

It takes around \textit{20} minutes for \textit{r}-NN graph construction and \textit{1.2} minutes to produce \textit{474,500} groups. In contrast, for the same task it would take more than \textit{100} hours for \textit{k}-means. Most of the clusters produced by our algorithms are meaningful. There are \textit{4,268} clusters that contain more than \textit{3} images. Since no ground-truth available, only three sample groups are shown in Fig.~\ref{fig:paris}(e). As shown in the figure, the algorithm performs reasonably well even with the support of approximate \textit{r}-NN graph. According to our observation, the small clusters (whose size is less than \textit{10}) are comprised by near-duplicate images, which is highly helpful for large-scale image linking tasks.

\section{Conclusion}
Boundary erosion is a process of unspinning the natural structures of potential clusters. The erosion starts from boundaries of clusters, invading inwards, till it reaches to all the density peaks. It therefore produces a sequential order that following which the clusters are naturally reconstructed. In the whole process, only one parameter, namely the radius of neighborhood $r$ is involved. The density peaks, the corresponding clusters and the cluster boundaries emerge automatically. The effectiveness of the algorithm has been verified on various clustering tasks and in different scales. Due to its simplicity, genericness, speed efficiency as well as superior performance across various datasets, this algorithm will find its value in various science and engineering tasks.

\section{Acknowledgments}
This work is supported by National Natural Science Foundation of China under grants 61572408.

{\small
\bibliographystyle{ieee}
\bibliography{cvpr}

\begin{thebibliography}{10}\itemsep=-1pt

\bibitem{mm16:amato}
G.~Amato, F.~Falchi, C.~Gennaro, and F.~Rabitti.
\newblock {YFCC100M} hybridnet fc6 deep features for content-based image
  retrieval.
\newblock In {\em The 2016 ACM Workshop on Multimedia COMMONS}, pages 11--18,
  2016.

\bibitem{optics99}
M.~Ankerst, M.~M. Breunig, H.-P. Kriegel, and J.~Sander.
\newblock {OPTICS}: ordering points to identify the clustering structure.
\newblock In {\em ACM Sigmod record}, pages 49--60. ACM, 1999.

\bibitem{peel17:nadav}
N.~Bar, H.~Averbuch{-}Elor, and D.~Cohen{-}Or.
\newblock Border-peeling clustering.
\newblock {\em CoRR}, abs/1612.04869, 2016.

\bibitem{pr08:chang}
H.~Chang and D.-Y. Yeung.
\newblock Robust path-based spectral clustering.
\newblock {\em Pattern Recognition}, 41(1):191--203, January 2008.

\bibitem{pathbased}
H.~Chang and D.-Y. Yeung.
\newblock Robust path-based spectral clustering.
\newblock {\em Pattern Recognition}, 41(1):191--203, 2008.

\bibitem{meanshift}
Y.~Cheng.
\newblock Mean shift, mode seeking, and clustering.
\newblock {\em IEEE Transactions on Pattern Analysis and Machine Intelligence},
  17(8):790--799, August 1995.

\bibitem{iccv17}
K.~G. Dizaji, A.~Herandi, C.~Deng, W.~Cai, and H.~Huang.
\newblock Deep clustering via joint convolutional autoencoder embedding and
  relative entropy minimization.
\newblock In {\em 2017 IEEE International Conference on Computer Vision
  (ICCV)}, pages 5747--5756. IEEE, 2017.

\bibitem{weidong}
W.~Dong, C.~Moses, and K.~Li.
\newblock Efficient k-nearest neighbor graph construction for generic
  similarity measures.
\newblock In {\em International Conference on World Wide Web}, pages 577--586,
  Mar. 2011.

\bibitem{markov}
S.~Dongen.
\newblock A cluster algorithm for graphs.
\newblock Technical report, CWI (Centre for Mathematics and Computer Science),
  Amsterdam, The Netherlands, 2000.

\bibitem{ssc}
E.~Elhamifar and R.~Vidal.
\newblock Sparse subspace clustering: Algorithm, theory, and applications.
\newblock {\em IEEE transactions on pattern analysis and machine intelligence},
  35(11):2765--2781, 2013.

\bibitem{dbscan}
M.~Ester, H.~peter Kriegel, J.~Sander, and X.~Xu.
\newblock A density-based algorithm for discovering clusters in large spatial
  databases with noise.
\newblock In {\em IEEE Transactions on Knowledge Discovery and Data
  Engineering}, pages 226--231, 1996.

\bibitem{pr06:franti}
P.~Fr\"{a}nti and O.~Virmajoki.
\newblock Iterative shrinking method for clustering problems.
\newblock {\em Pattern Recognition}, 39(5):761--775, May 2006.

\bibitem{sci07:frey}
B.~J. Frey and D.~Dueck.
\newblock Clustering by passing messages between data points.
\newblock {\em Science}, 315:972--976, Feburary 2007.

\bibitem{bmc07:fu}
L.~Fu and E.~Medico.
\newblock {FLAME}, a novel fuzzy clustering method for the analysis of dna
  microarray data.
\newblock {\em BMC Bioinformatics}, 8(3), January 2007.

\bibitem{dbscan15}
J.~Gan and Y.~Tao.
\newblock Dbscan revisited: Mis-claim, un-fixability, and approximation.
\newblock In {\em Proceedings of the 2015 ACM SIGMOD International Conference
  on Management of Data}, pages 519--530, 2015.

\bibitem{kdd07:gionis}
A.~Gionis, H.~Mannila, and P.~Tsaparas.
\newblock Clustering aggregation.
\newblock {\em ACM Transactions on Knowledge Discovery from Data}, 1(1), March
  2007.

\bibitem{ijirr13:greenlaw}
R.~Greenlaw and S.~Kantabutra.
\newblock Survey of clustering: Algorithms and applications.
\newblock {\em International Journal of Information Retrieval and Resouces},
  3(2):1--29, April 2013.

\bibitem{jain}
A.~K. Jain and M.~H. Law.
\newblock Data clustering: A user's dilemma.
\newblock {\em PReMI}, 3776:1--10, 2005.

\bibitem{JDS11}
H.~J\'egou, M.~Douze, and C.~Schmid.
\newblock Product quantization for nearest neighbor search.
\newblock {\em IEEE Transactions on Pattern Analysis and Machine Intelligence},
  33(1):117--128, Janurary 2011.

\bibitem{nips17:deepssc}
P.~Ji, T.~Zhang, H.~Li, M.~Salzmann, and I.~Reid.
\newblock Deep subspace clustering networks.
\newblock In {\em Advances in Neural Information Processing Systems}, pages
  23--32, 2017.

\bibitem{vade}
Z.~Jiang, Y.~Zheng, H.~Tan, B.~Tang, and H.~Zhou.
\newblock Variational deep embedding: An unsupervised and generative approach
  to clustering.
\newblock In {\em International Joint Conference on Artificial Intelligence},
  2017.

\bibitem{chameleon99}
G.~Karypis, E.-H.~S. Han, and V.~Kumar.
\newblock Chameleon: Hierarchical clustering using dynamic modeling.
\newblock {\em Computer}, 32(8):68--75, August 1999.

\bibitem{agnes90}
L.~Kaufman and P.~J. Rousseeuw.
\newblock {\em Finding Groups in Data: An Introduction to Cluster Analysis}.
\newblock John Wiley \& Sons, 1990.

\bibitem{density_review}
H.-P. Kriegel, P.~Kr{\"o}ger, J.~Sander, and A.~Zimek.
\newblock Density-based clustering.
\newblock {\em Wiley Interdisciplinary Reviews: Data Mining and Knowledge
  Discovery}, 1(3):231--240, 2011.

\bibitem{eyaleb}
K.-C. Lee, J.~Ho, and D.~J. Kriegman.
\newblock Acquiring linear subspaces for face recognition under variable
  lighting.
\newblock {\em IEEE Transactions on pattern analysis and machine intelligence},
  27(5):684--698, 2005.

\bibitem{km82}
S.~P. Lloyd.
\newblock Least squares quantization in {PCM}.
\newblock {\em IEEE Transactions on Information Theory}, 28:129--137, March
  1982.

\bibitem{pami14:flann}
M.~Muja and D.~G. Lowe.
\newblock Scalable nearest neighbor algorithms for high dimensional data.
\newblock {\em IEEE Transactions on Pattern Analysis and Machine Intelligence},
  36:2227--2240, 2014.

\bibitem{coil100}
S.~Nene, S.~Nayar, and H.~Murase.
\newblock Columbia object image library (coil 100).
\newblock {\em Technical Report CUCS-005-96}, 1988.

\bibitem{coil20}
S.~A. Nene, S.~K. Nayar, H.~Murase, et~al.
\newblock Columbia object image library (coil-20).
\newblock {\em Technical report CUCS-005-96}, 1996.

\bibitem{clusterone}
T.~Nepusz, H.~Yu, and A.~Paccanaro.
\newblock Detecting overlapping protein complexes in protein-protein
  interaction networks.
\newblock {\em Nature methods}, 9(5):471--472, 2012.

\bibitem{rank_order}
C.~Otto, D.~Wang, and A.~Jain.
\newblock Clustering millions of faces by identity.
\newblock {\em IEEE Transactions on Pattern Analysis and Machine Intelligence},
  2017.

\bibitem{decode09}
T.~Pei, A.~Jasra, D.~J. Hand, A.-X. Zhu, and C.~Zhou.
\newblock {DECODE}: a new method for discovering clusters of different
  densities in spatial data.
\newblock {\em Data Mining and Knowledge Discovery}, 18(3):337--369, 2009.

\bibitem{sci14:alex}
A.~Rodriguez and A.~Laio.
\newblock Clustering by fast search and find of density peaks.
\newblock {\em Science}, 344(6191):1492--1496, June 2014.

\bibitem{watershed00}
J.~B. Roerdink and A.~Meijster.
\newblock The watershed transform: Definitions, algorithms and parallelization
  strategies.
\newblock {\em Journal Fundamenta Informaticae}, 41(2):187--228, April 2000.

\bibitem{normcut}
J.~Shi and J.~Malik.
\newblock Normalized cuts and image segmentation.
\newblock {\em IEEE Transactions on Pattern Analysis and Machine Intelligence},
  22(8):888--905, August 2000.

\bibitem{SiZ03}
J.~Sivic and A.~Zisserman.
\newblock {Video Google}: {A} text retrieval approach to object matching in
  videos.
\newblock In {\em IEEE International Conference on Computer Vision}, pages
  1470--1477, October 2003.

\bibitem{face94}
F.~S. Smaria and A.~C. Harter.
\newblock Parameterisation of a stochastic model for human face identification.
\newblock In {\em IEEE Workshop on Applications of Computer Vision}, pages
  138--142, 1994.

\bibitem{hierarchical}
R.~C. Team.
\newblock R: A language and environment for statistical computing.
\newblock Technical report, R Foundation for Statistical Computing., 2012.

\bibitem{yfcc}
B.~Thomee, D.~A. Shamma, G.~Friedland, B.~Elizalde, K.~Ni, D.~Poland, D.~Borth,
  and L.-J. Li.
\newblock {YFCC100M}: The new data in multimedia research.
\newblock {\em Communications of the ACM}, 59(2):64--73, Feb. 2016.

\bibitem{transitive}
T.~Wittkop, D.~Emig, S.~Lange, S.~Rahmann, M.~Albrecht, J.~H. Morris,
  S.~B{\"o}cker, J.~Stoye, and J.~Baumbach.
\newblock Partitioning biological data with transitivity clustering.
\newblock {\em Nature methods}, 7(6):419--420, 2010.

\bibitem{nature15}
C.~Wiwie, J.~Baumbach, and R.~R{\"o}ttger.
\newblock Comparing the performance of biomedical clustering methods.
\newblock {\em Nature methods}, 12(11):1033--1038, 2015.

\bibitem{dec}
J.~Xie, R.~Girshick, and A.~Farhadi.
\newblock Unsupervised deep embedding for clustering analysis.
\newblock In {\em International Conference on Machine Learning}, pages
  478--487, 2016.

\bibitem{itnn05:xu}
R.~Xu and D.~I. Wunsch.
\newblock Survey of clustering algorithms.
\newblock {\em Transactions on Neural Networks}, 16(3):645--678, May 2005.

\bibitem{ocdt16:zdyu}
Z.~Yu, W.~Liu, W.~Liu, Y.~Yang, M.~Li, and B.~V. K.~V. Kumar.
\newblock On order-constrained transitive distance clustering.
\newblock In {\em Proceedings of the Thirtieth AAAI Conference on Artificial
  Intelligence}, AAAI'16, pages 2293--2299. AAAI Press, 2016.

\bibitem{ml04:zhao}
Y.~Zhao and G.~Karypis.
\newblock Empirical and theoretical comparisons of selected criterion functions
  for document clustering.
\newblock {\em Machine Learning}, 55:311--331, 2004.

\bibitem{ensemble14:zheng}
L.~Zheng, T.~Li, and C.~Ding.
\newblock A framework for hierarchical ensemble clustering.
\newblock {\em ACM Transactions on Knowledge Discovery from Data},
  9(2):9:1--9:23, September 2014.

\bibitem{ensemble12:zhou}
Z.-H. Zhou.
\newblock {\em Ensemble Methods: Foundations and Algorithms}.
\newblock Chapman \& Hall/CRC, 1st edition, 2012.

\end{thebibliography}
}

\end{document}